\documentclass{ieeeaccess}
\usepackage{cite}
\usepackage{amsmath,amssymb,amsfonts}
\usepackage{graphicx}
\usepackage{textcomp}
\usepackage{comment}
\usepackage[ruled,vlined]{algorithm2e}
\usepackage{caption}
\usepackage{subcaption}
\usepackage[hidelinks]{hyperref}
\usepackage{multirow}

\usepackage{bm}
\makeatletter
\AtBeginDocument{\DeclareMathVersion{bold}
\SetSymbolFont{operators}{bold}{T1}{times}{b}{n}
\SetSymbolFont{NewLetters}{bold}{T1}{times}{b}{it}
\SetMathAlphabet{\mathrm}{bold}{T1}{times}{b}{n}
\SetMathAlphabet{\mathit}{bold}{T1}{times}{b}{it}
\SetMathAlphabet{\mathbf}{bold}{T1}{times}{b}{n}
\SetMathAlphabet{\mathtt}{bold}{OT1}{pcr}{b}{n}
\SetSymbolFont{symbols}{bold}{OMS}{cmsy}{b}{n}
\renewcommand\boldmath{\@nomath\boldmath\mathversion{bold}}}
\makeatother

\def\BibTeX{{\rm B\kern-.05em{\sc i\kern-.025em b}\kern-.08em
    T\kern-.1667em\lower.7ex\hbox{E}\kern-.125emX}}

\begin{document}
\history{Date of publication xxxx 00, 0000, date of current version xxxx 00, 0000.}
\doi{10.1109/ACCESS.2024.0429000}

\title{Clustering-Based Collective Anomaly
Detection in IoT Systems: A Graph
Neural Network Approach}
\author{\uppercase{Dalila Khettaf}\authorrefmark{1}, 
\uppercase{Djamel Djenouri}\authorrefmark{1}, 
\uppercase{Zeinab Rezaeifar}\authorrefmark{1},
AND \uppercase{Youcef Djenouri}\authorrefmark{2},
}

\address[1]{University of the West of England(UWE) Bristol BS16 1QY, England (e-mail: dalila.khettaf@uwe.ac.uk)}
\address[2]{University of South-Eastern Norway, Kongsberg, 3616 Norway(e-mail: youcef.djenouri@usn.no)}


\markboth
{Author \headeretal: Preparation of Papers for IEEE TRANSACTIONS and JOURNALS}
{Author \headeretal: Preparation of Papers for IEEE TRANSACTIONS and JOURNALS}

\corresp{Corresponding author: Dalila Khettaf (e-mail: dalila.khettaf@uwe.ac.uk).}

\begin{abstract}
The rapid advancement of Internet of Things (IoT) technology has led to the widespread deployment of smart, interconnected devices across a range of domains. However, this expansion has also resulted in a substantial increase in network traffic, creating more opportunities for malicious actors to launch cyberattacks and compromise sensitive information, thereby increasing the need for effective anomaly detection. The state-of-the-art in anomaly detection has predominantly focused on point anomalies. In contrast, the detection
of collective anomalies remains relatively under-explored in the literature. 

In this paper, we introduce Unsupervised Graph
Collective Anomaly Detection (UGCAD), a novel framework
designed to identify collective anomalies in IoT network traffic.
Unlike many existing methods, UGCAD operates on graph-structured data without any prior knowledge of group labels
or membership. It leverages a variational graph autoencoder
(VGAE) to learn the graph representation, which is subsequently
used to enhance a clustering algorithm for effective grouping
of nodes. To detect collective anomalies, clusters identified as
normal are first aggregated and refined, after which anomaly
scores are applied to detect collective anomalies. Extensive experiments conducted on the CICIoT2023 and ToN-IoT network datasets demonstrate the
effectiveness of UGCAD in both clustering and collective anomaly detection (CAD). Furthermore, comparative evaluations against several traditional and state-of-the-art clustering-based CAD approaches confirm the superiority of UGCAD in accurately detecting collective
anomalies.

\end{abstract}

\begin{keywords}

Collective anomaly detection, deep learning, graph neural networks, internet of things, variational graph auto-encoder.

\end{keywords}

\titlepgskip=-21pt

\maketitle

\section{Introduction}
\label{sec:introduction}

\IEEEPARstart{T}{he} integration and widespread adoption of Internet of Things (IoT) technology have led to the proliferation of IoT-enabled smart devices across a wide range of applications, including smart home appliances (e.g., smart TVs, thermostats, smart meters, and security systems), healthcare devices, wearable devices, and vehicular and roadside systems \cite{qin2024partially}. These devices enable a plethora of applications but operate on diverse technologies and communication protocols, often forming a heterogeneous network. Combined with the large volumes of data they generate and manage, this complexity increases the vulnerability to cyberattacks, placing users’ sensitive information at risk \cite{javeed2023intelligent} \cite{ozadowicz2018application}.

Anomaly detection is a fundamental approach for mitigating cyberattacks and ensuring the security of IoT networks. It involves identifying observations or patterns in data that deviate markedly from the expected behavior of a system \cite{pang2021deep}. This capability is critical for ensuring the dependable and trustworthy operation of complex systems across a wide range of domains, including healthcare, cybersecurity, finance, and e-commerce \cite{pang2021deep}. In the literature, anomalies are generally categorized into three types: point anomalies, contextual anomalies, and collective anomalies \cite{chandola2009anomaly}. 
A point anomaly refers to a single data instance that deviates significantly from the expected behavior of a system. A contextual (or conditional) anomaly is a data instance that is considered anomalous only within a specific context. In contrast, a collective (or group) anomaly consists of a set of data instances that together exhibit anomalous behavior, even if individual instances appear normal \cite{Khettaf2026deep}. A common example of a collective anomaly is the Denial of Service (DoS) attack, along with its distributed version (DDoS), in which a server or network is overwhelmed by a flood of requests. While a single request appears legitimate, the overall pattern of traffic reveals the anomalous nature of the attack \cite{ahmed2018collective}. Another example is the Mirai attack, a large-scale cyber incident in 2016 in which malware infected insecure IoT devices, turning them into a botnet. This botnet was then used to launch powerful DDoS attacks, overwhelming major online platforms and services.

While point anomaly detection has been extensively investigated in the literature with methods such as Local Outlier Factor (LOF) \cite{breunig2000lof}, Isolation Forest (IF) \cite{liu2008isolation}, k-Nearest Neighbors (KNN) \cite{peterson2009k}, and One-Class Support Vector Machine (OCSVM) \cite{shin2005one}, research on collective anomaly detection (CAD) has been comparatively limited. Initial studies in this area predominantly relied on statistical and probabilistic methods \cite{yu2015glad} \cite{xiong2011group}. More recent advancements, however, have increasingly incorporated deep learning (DL) techniques, with a particular emphasis on applications involving time series data \cite{li2022unified} \cite{weng2019collective} \cite{bontemps2016collective}.

Graphs effectively represent relationships between entities and can model diverse data scenarios, notably when combined with DL, i.e., Graph Neural Networks (GNNs), which perform well on both node and graph-level tasks such as classification and anomaly detection. However, graph data remains underexplored in CAD, with most existing approaches assuming group membership is known in advance \cite{van2024towards}. Motivated by these challenges, we propose a clustering-based approach that leverages GNNs to identify collective anomalies. In particular, our method employs a variational graph auto-encoder (VGAE) to learn embeddings from IoT network traffic represented as a graph. These embeddings are subsequently grouped using a clustering algorithm, and the resulting clusters are analyzed with an anomaly detection technique to uncover collective anomalies in the data. The main contributions of this article are:
\begin{itemize}
    \item Introducing a CAD approach that leverages clustering and node embeddings to detect network attacks, without requiring prior knowledge of group membership.
    \item Evaluating the proposed method on real-world datasets, testing its scalability and robustness with different graph sizes and different anomaly ratios, and comparing its performance against state-of-the-art approaches.
    \item Demonstrating the broader applicability of CAD for network traffic analysis.
\end{itemize}

The remainder of this paper is organized as follows. Section \ref{sec:related-work} discusses the related work. Section \ref{sec:pf} formulates the problem of CAD. Section \ref{sec:sm} presents the system model for UGCAD. Section \ref{sec:ps} elaborates the proposed solution. Section \ref{sec:exp} provides the experimental details and Section \ref{sec:pe} presents the results. Finally, Section \ref{sec: conc} concludes the paper.

\section{Related Work} \label{sec:related-work}
Earlier research on CAD primarily relied on statistical and probabilistic approaches. Yu et al. \cite{yu2015glad} introduced a hierarchical Bayesian model to identify group anomalies in social media data. Xiong et al. \cite{xiong2011group} presented  an extension of LDA (Latent Dirichlet allocation) to detect collective anomalies in predefined groups.
Song et al. \cite{song2020group} proposed a method that captures correlations between groups and built a full variational Bayesian framework that integrates a genetic algorithm for parameter optimization. 
These statistical methods rely on strong prior assumptions (e.g., predefined groups or distributions) and detect anomalies based on statistical properties like variance or likelihood shifts. As a result, they often fail when anomalies are subtle or do not significantly deviate statistically, and they struggle to capture complex relational structures in the data.

More recent approaches to CAD increasingly incorporate advanced machine learning algorithms and extend to diverse data modalities, including images, time series, and graphs. In particular, numerous methods have been proposed for detecting collective anomalies in time series data. For instance,
Li et al. \cite{li2022unified} introduced Stacked Temporal Convolutional Networks (CPA-TCN) for detecting both point and collective anomalies. Weng et al. \cite{weng2019collective} introduced a method that leverages statistical features together with the isolation forest algorithm to detect collective anomalies in multidimensional data streams within cloud environments. Bontemps et al. \cite{bontemps2016collective} adapted a Long Short-Term Memory (LSTM) model to detect collective anomalies in time series. Shi et al.\cite{shi2023bidirectional} modeled time series using linear fitting functions and then applied Piecewise Integration (PI) to detect collective anomalies in both synthetic and real-world datasets.
Other approaches have been developed for image data. For example, Belhadi et al. \cite{belhadi2021deep} investigated both data mining techniques and Convolutional Neural Networks (CNNs) for detecting collective abnormal human behavior. Chalapathy et al.\cite{chalapathy2018group} proposed to use both Adversarial Autoencoders (AAEs) and Variational Autoencoders (VAEs) for detecting collective anomalies in images, treating each image as a group of pixels.

Relatively few studies have focused on CAD for graph data. For instance, Feng et al. \cite{feng2022full} first converted multivariate time series into graphs and then proposed both a Graph Autoencoder (GAE) and a VGAE to detect collective anomalies in the Industrial Internet of Things (IIoT) systems. D’Oro et al. \cite{d2019group} presented a method that utilizes graph representation learning to detect collective anomalies. Ai et al. \cite{ai2024graph} proposed a framework that leverages a GAE and anchor nodes to identify anomalous groups.

These recent DL-based approaches remain relatively under-developed. Many methods still rely on prior knowledge or predefined assumptions, limiting their generalizability. Time series-based approaches frequently employ sliding window mechanisms, which are inherently constrained and may fail to capture anomalies that do not conform to fixed window boundaries. Image-based methods are also limited, as collective anomalies in visual data are not well-established and are often evaluated on simplified or synthetic scenarios. Similarly, graph-based approaches often depend on predefined structures or unrealistic settings, restricting their effectiveness in modeling complex and real-world collective anomaly patterns.

Other approaches adopted a different strategy, identifying anomalous groups by first clustering the data. For example, Mirsky et al. \cite{mirsky2017anomaly} proposed a clustering approach based on Principal Component Analysis (PCA) to detect group anomalies in smartphone data streams. 
The authors of \cite{ahmed2015novel} introduced MCADET, an approach that relies on X-Means clustering and cluster variance to detect flooding-based DoS attacks.
Belhadi et al. \cite{belhadi2021hybrid} proposed HDM-GAD, a hybrid framework that combines DBSCAN for micro-clustering with a pruning strategy to detect collective anomalies in sequence data. Ahmed et al. \cite{ahmed2018collective} proposed HCADET, a method for detecting DoS attacks in network traffic data that uses two-stage X‑Means clustering and the Hurst parameter. 
Wang et al. \cite{wang2022network} proposed CCAD, a framework that detects collective anomalies in streaming network traffic by tracking changes in clustering structure within sliding windows. The authors of \cite{ahmed2014network} proposed CADET, a method that relies on theoretic co-clustering to detect collective anomalies in network traffic data.  

Although clustering-based methods provide an initial step toward grouping collective anomalies, they rely on overly simplistic assumptions to distinguish normal from anomalous clusters, typically based on basic properties such as cluster size or statistical features.

Table \ref{tab:cad_comparison} presents a comparative analysis of the aforementioned methods based on several important characteristics. Specifically, the table indicates whether each method: (1) employs DL techniques, (2) operates on graph-structured data, (3) detects collective anomalies, (4) uses clustering algorithms to group data, (5) relies on prior knowledge of group memberships, and (6) follows a supervised learning paradigm. As shown in the table, our proposed method, UGCAD, combines several desirable properties by leveraging DL on graph data for CAD while incorporating clustering mechanisms without requiring prior knowledge of group memberships. Furthermore, UGCAD operates in an unsupervised manner, making it more suitable for realistic scenarios where labeled data and predefined groups are unavailable.

Overall, existing methods for CAD rely on strong assumptions, prior knowledge, or simplistic statistical properties, limiting their ability to capture complex and subtle anomaly patterns. Moreover, many approaches fail to model realistic structures and dependencies in data, resulting in poor generalization to real-world and dynamic scenarios. 
In this work, we propose UGCAD, a fully unsupervised clustering-based CAD method for graph data that requires no prior knowledge of group membership, avoids excessive hyperparameters, and makes no assumptions about the underlying data.

\begin{table}[ht]
\centering
\caption{Comparison of Collective Anomaly Detection Methods}
\label{tab:cad_comparison}

\resizebox{\linewidth}{!}{%
\renewcommand{\arraystretch}{1.2}
\begin{tabular}{|l|c|c|c|c|c|c|}
\hline
\textbf{Method} & \textbf{DL} & \textbf{Clustering} & \textbf{Graph Data} & \textbf{CAD} & \textbf{Prior Knowledge} & \textbf{Supervised} \\
\hline
GLAD \cite{yu2015glad } & No & No  & Yes & Yes & No  & No  \\
\hline
FGM \cite{xiong2011group} & No  & Yes & Yes  & No & No & No \\
\hline
CHGM \cite{song2020group} & No & No & Yes & No  & Yes & No  \\
\hline
CPA-TCN \cite{li2022unified} & Yes  & No  & Yes & No & No  & No \\
\hline
iForestFS \cite{weng2019collective}  & No & No  & Yes  & No & No & No  \\
\hline
Method by \cite{bontemps2016collective} & Yes & No  & Yes  & No & No & No  \\
\hline
BPLR \cite{shi2023bidirectional}& No & No  & Yes  & No & No & No  \\
\hline
CaHB \cite{belhadi2021deep} & Yes & No  & Yes  & Partially & No & Mixed  \\
\hline
AAE/VAE \cite{chalapathy2018group} & Yes & No  & Yes  & No & Yes & No  \\
\hline
f-GAE/f-VGAE \cite{feng2022full}& Yes & Yes  & Yes  & No & Yes & No  \\
\hline
GADGA \cite{d2019group} & Yes & Yes  & Yes  & No & Yes & No  \\
\hline
TP-GrGAD \cite{ai2024graph} & Yes & Yes  & Yes  & No & No & No  \\
\hline
pcStream \cite{mirsky2017anomaly} & No & No  & Yes  & Yes & No & No  \\
\hline
HDM-GAD \cite{belhadi2021hybrid} & No & No  & Yes  & Yes & No & No  \\
\hline
CCAD \cite{wang2022network} & No & No  & Yes  & Yes & No & No  \\
\hline
CADET/MCADET/HCADET \cite{ahmed2015novel} & No & No  & Yes  & Yes & No & No  \\
\hline
Ours(UGCAD) & Yes & Yes  & Yes  & Yes & No & No  \\
\hline
\end{tabular}%
}
\end{table}

\section{Problem Formulation} \label{sec:pf}

The problem of collective (or group) anomaly detection can be defined as a function that takes data as input and outputs a set of anomalous groups. In this work, we adapt the CAD problem to graph data.

Let $G = (V, E)$ denote an undirected graph, where 
$V = \{v_i\}_{i=1}^{N}$ is the set of nodes and 
$E = \{  e_{i, j} \mid \{i, j\} \subseteq V \}$ is the set of edges. The relationships between nodes can be represented by an adjacency matrix $A \in \{0,1\}^{N \times N}$ described in Eq. \ref{eq:adj}:

\begin{equation}
\label{eq:adj}
A_{ij} = 
\begin{cases}
1, & \text{if } e_{i,j} \in E, \\
0, & \text{otherwise},
\end{cases}
\quad \forall i,j \in V.
\end{equation}

Each node $v \in V$ is associated with attributes represented by a feature matrix $X \in \mathbb{R}^{N \times d}$, where $d$ is the dimensionality of node features.

We propose an \textit{Unsupervised Graph-based Collective Anomaly Detection} (UGCAD) method, formalized as a function $F : (A, X) \to \mathcal{D}$, where $\mathcal{C} = \{c_1, c_2, \dots, c_m\}$ denotes the set of all candidate clusters, and $\mathcal{D} \subseteq \mathcal{C}
$ represents the subset of anomalous clusters or collective anomalies. Each cluster $c \in \mathcal{C}$ is assigned an anomaly score by a scoring function $ S : \mathcal{C} \to \mathbb{R}$.

A cluster $c \in \mathcal{C}$ is considered anomalous if its score exceeds a threshold $\tau$, formulated in Eq. \eqref{eq:cind}:

\begin{equation}
\label{eq:cind}
    c \in \mathcal{D} \iff S(c) > \tau
\end{equation}

\section{System Model} \label{sec:sm}

The main purpose of the proposed method, UGCAD, is to detect collective anomalies in network traffic and produce a set of normal clusters and a set of anomalous clusters. The solution can be integrated into an IoT system and works with various IoT devices, including smart home hubs, industrial sensors, smart cameras, wearables, and so on.

UGCAD, originally designed for detecting collective anomalies in graphs, can be adapted for streaming IoT network data using clustering algorithms such as DenStream \cite{cao2006density}. DenStream is used here as an example, but any suitable online clustering algorithm could be applied in its place. DenStream is an online density-based clustering algorithm designed for data streams; it maintains compact summaries called micro-clusters that represent dense regions (normal patterns) while tracking sparse or emerging regions as potential anomalies. It continuously updates these micro-clusters as new data arrives and applies decay to outdated patterns, making it suitable for evolving IoT environments.
To reduce the computational load on IoT devices, UGCAD is first pretrained in the cloud, where clusters representing normal and anomalous behavior are generated. Normal clusters are converted into Core Micro-Clusters (CMCs), while anomalous clusters become Outlier Micro-Clusters (OMCs). These are then deployed to IoT edge gateways or local servers.

As new network traffic arrives, it is matched to the nearest micro-cluster. If it fits within a CMC, it reinforces normal behavior; if it fits an OMC or leads to the creation of a new outlier cluster, it is flagged as potentially anomalous. Outlier clusters that accumulate enough weight can be promoted to CMCs to capture evolving normal patterns, while existing clusters decay over time to handle concept drift.

This architecture enables UGCAD to detect collective anomalies in real time, even when individual network packets appear normal. Heavy computation is offloaded to the cloud, while IoT devices and edge gateways perform lightweight online processing. Periodically, micro-cluster summaries and detected anomalies are sent back to the cloud for monitoring, visualization, and retraining.
\section{Proposed Solution} \label{sec:ps}

\begin{figure*}[ht!]
    \centering
    \includegraphics[width=0.7\textwidth, height = 0.3\textheight]{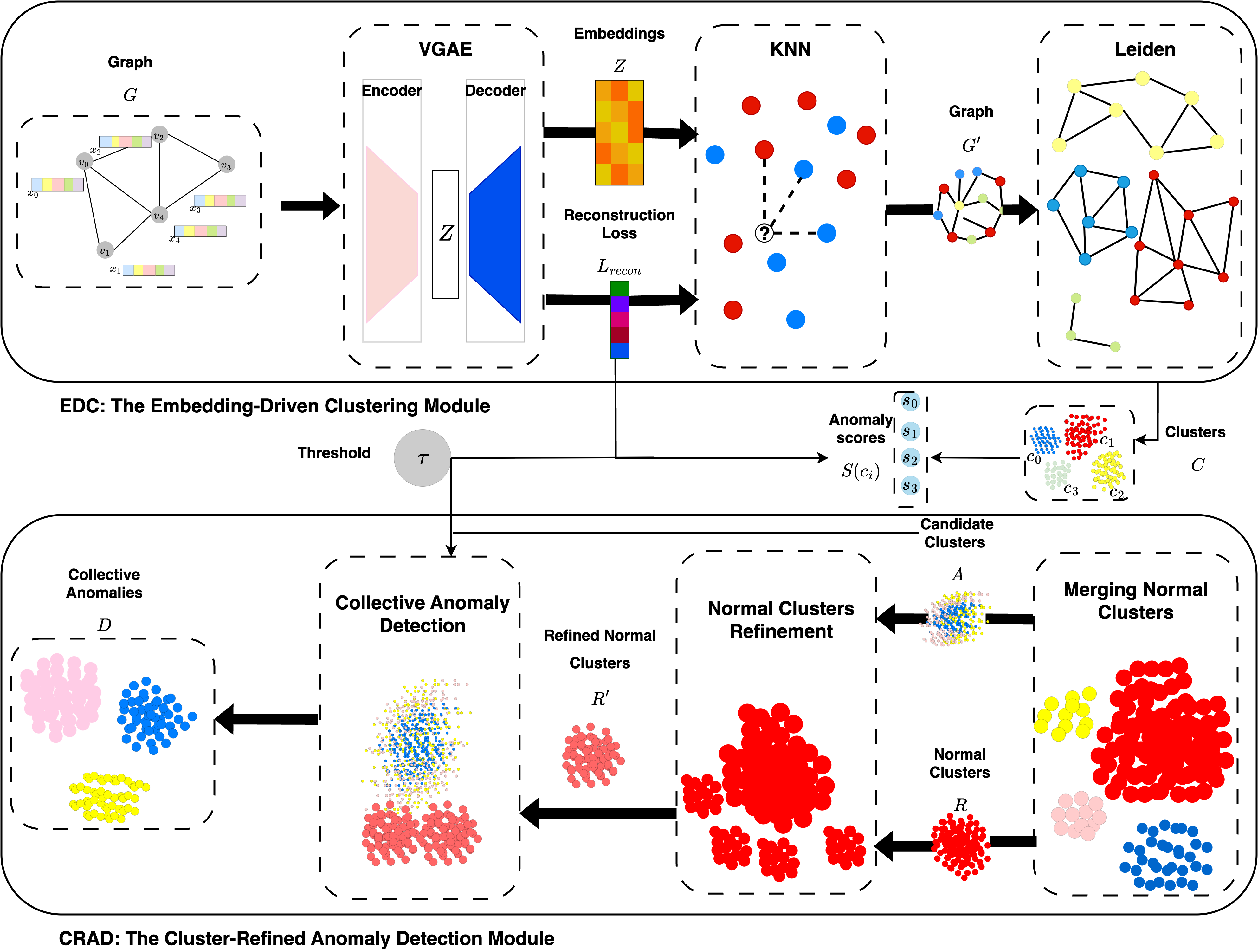}
    \caption{The Proposed UGCAD Approach.}
    \label{fig:UGCAD}
\end{figure*}

The proposed method, UGCAD, consists of two modules as illustrated in Fig. \ref{fig:UGCAD}. The first is the Embedding-Driven Clustering (EDC) module, which applies the Leiden algorithm to node embeddings generated by a VGAE, thereby capturing the underlying community structure of the graph. The second is the Cluster-Refined Anomaly Detection (CRAD) module, which leverages these clustering results to identify collective anomalies by distinguishing anomalous cluster formations from the refined normal ones. 

A collective anomaly is, by definition, a group of nodes that exhibit anomalous behavior only when considered together. This motivates the first step of our method: clustering similar nodes, which captures the collective aspect of the anomaly. The second step is to determine which clusters are anomalous. To do this, we rely on a simple intuition grounded in reconstruction-based anomaly detection. In a collective anomaly, some nodes will individually appear abnormal, and a VGAE will fail to reconstruct them, yielding high reconstruction errors, a standard anomaly indicator in the literature of graph anomaly detection. However, as discussed earlier, not all nodes within a collective anomaly are necessarily anomalous on their own. Therefore, we compute the anomaly score of a cluster by averaging the reconstruction errors of its nodes, which helps smooth out the influence of nodes that are not individually anomalous while still capturing the collective abnormality.
The rest of this section details the two modules.

\subsection{The Embedding-Driven Clustering (EDC) Module}
EDC takes the input graph $G$ and produces a set of clusters $C$, as summarized in Algorithm \ref{algo:EDC}. The primary objective of this module is to separate collective anomalies from normal data by grouping them into distinct clusters. This design is motivated by our previous work \cite{khettaf2025hybrid}, that proposed a community detection framework that enhances the Louvain algorithm \cite{blondel2008fast} with node embeddings generated using a GNN. Authors of \cite{khettaf2025hybrid} also introduced a merging procedure to consolidate highly similar clusters and demonstrated that using learned node embeddings rather than raw features as input to Louvain results in clusters of significantly higher quality.

GAEs are an extension of traditional autoencoders(AEs) to graph-structured data, where the goal is to learn low-dimensional node embeddings that capture the graph’s structure. VGAEs further extend GAEs by modeling the latent embeddings probabilistically. VGAEs can be used to learn meaningful representations of nodes or entire graphs.

The proposed architecture uses a  Graph Convolutional Network (GCN) encoder. More formally, given node feature matrix $X$ and adjacency matrix $A$, the encoder applies two GCN layers with ReLU activations and dropout as shown in Eq. \ref{eq:hiddenlayers}:

\begin{equation}
\begin{aligned}
H^{(1)} = ReLU(GCNConv_1(X, A)) \\
H^{(2)} = ReLU(GCNConv_2(H^{(1)}, A))
\end{aligned}
\label{eq:hiddenlayers}
\end{equation}

From $H^{(2)}$, two separate GCN layers produce the mean $\mu$ and log-variance $\log \sigma^2$ for each node’s latent embedding, formulated in Eq. \ref{eq:muvar}:

\begin{equation}
\begin{aligned}
\mu = GCNConv_\mu(H^{(2)}, A) \\
\log \sigma^2 = GCNConv_{\log\text{var}}(H^{(2)}, A)
    \end{aligned}
    \label{eq:muvar}
\end{equation}

Latent embeddings are then sampled using the reparameterization trick, as described in Eq. \ref{eq:z}:

\begin{equation}
    z = \mu + \sigma \odot \epsilon, \quad \epsilon \sim \mathcal{N}(0, I)
    \label{eq:z}
\end{equation}

The graph is reconstructed implicitly through an inner product decoder $\hat{A} = \sigma(Z Z^T)$, where $\sigma$ is the sigmoid function. The reconstruction loss used is the binary cross-entropy loss computed over the set of positive edges $E^+$ and a set of negatively sampled edges $E^-$. The decoder models the probability of an edge between nodes $i$ and $j$ as $\hat{A}_{ij} = \sigma(z_i^\top z_j)$ and the loss $\mathcal{L}_{recon}$ minimizes the negative log-likelihood of the observed edges and sampled non-edges, formulated in Eq. \ref{eq:Lrecon}:

\begin{equation}
    \mathcal{L}_{\text{recon}} = - \frac{1}{|E^+|} \sum_{(i,j)\in E^+} \log \hat{A}_{ij} 
- \frac{1}{|E^-|} \sum_{(i,j)\in E^-} \log (1 - \hat{A}_{ij})
\label{eq:Lrecon}
\end{equation}

This objective encourages embeddings of connected nodes to yield high predicted probabilities, while embeddings of unconnected nodes are encouraged to yield low probabilities. In this way, the learned latent space captures and preserves the connectivity patterns of the input graph, producing meaningful probabilistic node representations.

\begin{algorithm}[t]

\caption{EDC: Embedding-Driven Clustering}

\label{algo:EDC}

\KwIn{Graph $G(A, X)$}

\KwOut{$C$, $Z$, $\mathcal{L}_{\text{recon}}$}

$(Z, \mathcal{L}_{\text{recon}}) \gets \text{VGAE}(G)$\
\\
$G' \gets \text{KNN}(Z)$\
\\
$\mathcal{C} \gets \text{Leiden}(G')$\

Return $\mathcal{C}, Z, \mathcal{L}_{\text{rec}}$
\end{algorithm}
The Louvain algorithm is an algorithm for community detection that iteratively maximizes modularity to cluster nodes into densely connected groups. However, unlike the work \cite{khettaf2025hybrid} that relies on the Louvain algorithm, we replace it with the Leiden algorithm \cite{traag2019louvain}. The Leiden algorithm was introduced as an improvement over Louvain after it was shown that Louvain can produce poorly connected communities and suffers from scalability issues \cite{traag2019louvain}. In contrast, Leiden guarantees well-connected communities, improves the quality of detected clusters, and is computationally faster, making it more suitable for large-scale graphs.
To generate a graph from the embeddings, we employ the k-nearest neighbors (k-NN) method as shown in Algorithm \ref{algo:EDC}, which is simple to implement and requires only a few hyperparameters. The k-NN algorithm connects each node to its closest neighbors with edges, resulting in a graph $G'$. $G'$ is then passed to the Leiden algorithm to identify the set of clusters $\mathcal{C}$. Finally, the cluster set $\mathcal{C}$, the embeddings $Z$, and the reconstruction loss $\mathcal{L}_{\text{recon}}$ are provided as input to the subsequent module.
\subsection{Cluster-Refined Anomaly Detection (CRAD) Module:}
The CRAD module is responsible for the detection of collective anomalies. It takes as input the set of clusters $\mathcal{C}$, node embeddings $Z$, and the reconstruction loss $\mathcal{L}_{\text{recon}}$. It outputs a set of collective anomalies $\mathcal{D}$. The steps of the algorithm are summarized in Algorithm
\ref{alg:crad}. 

The first step is to assign an anomaly score to each cluster. Formally, for each cluster $c_k \in \mathcal{C}$, the cluster-level anomaly score is defined using a scoring function $S$ as the average reconstruction error of its constituent nodes, shown in Eq. \ref{eq: score_cluster}:

\begin{equation}
\label{eq: score_cluster}
S(c_k) = \frac{1}{|c_k|} \sum_{i \in c_k} L_{\text{recon}}(z_i)
\end{equation}

The second step is to separate the large group of normal clusters. This is achieved using the merging algorithm from \cite{khettaf2025hybrid}, which relies on the pairwise cosine distance between clusters in the embedding space. Clusters that are close in terms of distance are selected, based on the intuition that the majority of normal clusters tend to be close together in the embedding space. As a result, the clusters are partitioned into a set of normal clusters \(\mathcal{R} \subseteq \mathcal{C}\) and a set of candidate anomalous clusters $\mathcal{A} = \mathcal{C} \setminus \mathcal{R}$. The candidate set $\mathcal{A}$ is not immediately classified as anomalous, since it may still contain some normal clusters, and there is no definitive proof that its members are truly anomalous. This cautious approach is necessary because collective anomalies can sometimes resemble normal data, meaning that normal clusters may remain within $\mathcal{A}$.
The third step involves refining the set of normal clusters $\mathcal{R}$. This refinement is crucial for accurately determining the anomaly detection threshold. Since the anomaly score is based on reconstruction loss, some normal clusters that resemble collective anomalies may exhibit relatively high reconstruction errors. To address this, a percentile threshold $T_\alpha$ is used, representing a small percentile ($\alpha$) of the highest cluster anomaly scores in the candidate set $\mathcal{A}$, formulated in Eq. \ref{eq:taualpha}:
\begin{equation}
    T_\alpha = \text{percentile}_\alpha \big( \{ S(c) \mid c \in \mathcal{A} \} \big)
    \label{eq:taualpha}
\end{equation}
Using this threshold, a refined set of normal clusters $R'$ is created by selecting all clusters in $\mathcal{R}$ whose anomaly scores are below the $T_\alpha$ threshold as described in Eq. \ref{eq:percentile}:
\begin{equation}
 \label{eq:percentile}   
\mathcal{R}' = \{ c \in \mathcal{R} \mid S(c) < T_\alpha \}
\end{equation}
This ensures that $\mathcal{R'}$ contains only the most confidently normal clusters, which in turn allows for a more reliable calculation of the anomaly detection threshold.

The fourth step consists of calculating the anomaly detection threshold $\tau$ to detect collective anomalies. 
Given a set of anomaly scores from normal clusters in $\mathcal{R'} : \{s_1, s_2, \dots, s_n\}$, 
the probability density function of the normal data is estimated using Kernel Density Estimation (KDE) in Eq. \ref{eq:kde}:
\begin{equation}
\hat{f}(s) = \frac{1}{n h} \sum_{i=1}^{n} K\left(\frac{s - s_i}{h}\right)
\label{eq:kde}
\end{equation}
where $K$ is the kernel function, typically chosen as a Gaussian, and $h$ is the bandwidth controlling the smoothness of the estimate. 
Both $K$ and $h$ are hyperparameters selected based on prior work \cite{feng2021make}.
The cumulative distribution function (CDF) is computed from the estimated density as (Eq.\ref{eq:cdf}) :
\begin{equation}
\hat{F}(s) = \int_{-\infty}^{s} \hat{f}(t) \, dt
\label{eq:cdf}
\end{equation}
which represents the proportion of normal observations below a given score $s$. 
The anomaly threshold $\tau$ is defined as the smallest value satisfying $\hat{F}(\tau) \geq \beta$, where $\beta$ is the confidence level chosen according to \cite{feng2021make} at 95\%. 
In practice, $\tau$ corresponds to the score below which approximately 95\% of normal clusters lie, 
and any cluster with a score exceeding $\tau$ is considered anomalous. 

In the last step, collective anomalies are identified from $\mathcal{A}$ by evaluating the anomaly scores of each cluster. 
A cluster $c_k \in \mathcal{A}$ is considered a collective anomaly if its score exceeds the previously calculated threshold $\tau$: $S(c_k) > \tau$, Thus, the set of collective anomalies is formally defined as $\mathcal{D} = \{ c \in \mathcal{A} \mid S(c) > \tau \}$.
This approach ensures that clusters with scores significantly higher than the threshold, representing deviations from normal behavior, are flagged as collective anomalies.

\begin{algorithm}[t]
\caption{CRAD: Cluster-Refined Anomaly Detection}
\label{alg:crad}
\KwIn{$\mathcal{C}$, $Z$, $\mathcal{L}_{\text{recon}}$}
\KwOut{ $\mathcal{D}$}

\BlankLine
\textbf{Step 1: Compute cluster-level anomaly scores} \\
\ForEach{$c_k \in \mathcal{C}$}{

Calculate $S(c_k)$ using Eq. \ref{eq: score_cluster}
}

\BlankLine
\textbf{Step 2: Separate normal and candidate anomalous clusters} \\
$\mathcal{R}\gets \text{MergeNormalClusters}(\mathcal{C}, Z)$ 
\\
$\mathcal{A} = \mathcal{C} \setminus \mathcal{R}$
\BlankLine

\textbf{Step 3: Refine normal clusters} \\
Calculate $T_\alpha$ using Eq. \ref{eq:taualpha} \\
Derive $\mathcal{R}'$ from Eq. \ref{eq:percentile} 
\BlankLine

\textbf{Step 4: Compute anomaly detection threshold} \\
Compute $\hat{F}$ using Eq. \ref{eq:kde} and Eq. \ref{eq:cdf}\\
$\tau \gets \min \{ s \mid \hat{F}(s) \geq \beta \}$ 
\BlankLine

\textbf{Step 5: Identify collective anomalies} \\
$\mathcal{D} \gets \{ c \in \mathcal{A} \mid S(c) > \tau \}$

\BlankLine
Return $\mathcal{D}$
\end{algorithm}

\section{Experiments} \label{sec:exp}

\subsection{Experimental Setup:}
The proposed method was implemented in Python using libraries including PyTorch, Pandas, and NumPy. Computationally intensive operations were accelerated using GPU processing. The experiments were conducted on a MacBook Air M1 and Google Colab. 

\subsection{Dataset}
To evaluate the proposed method, we utilized the CICIoT2023 dataset \cite{neto2023ciciot2023} and the network ToN-IoT dataset \cite{toniot_dataset}. The CICIoT2023 dataset is a real-world dataset collected by the Canadian Institute for Cybersecurity. This dataset comprises network traffic from $105$ IoT devices and contains $33$ different types of attacks. The attacks are categorized into the following classes: DDoS, Mirai, Reconnaissance, Brute Force, Spoofing, DoS, and Web-based attacks. The ToN-IoT network dataset, developed by the Cyber Range Lab at UNSW Canberra, contains network traffic from simulated IoT and IIoT environments. It includes both normal and malicious network flows collected in a realistic smart setting using tools like Argus and Zeek. As part of the ToN\_IoT collection, it supports evaluating AI-based cybersecurity methods and contains cyber-attacks such as DoS, DDoS, and scanning.
\hspace{-100pt}
\subsection{Dataset Pre-Processing}
For dataset preprocessing, irrelevant features were removed, non-numerical values were converted to numerical ones, and the resulting data were standardized using StandardScaler from scikit-learn.
From the original CICIoT2023 dataset, we constructed a subset comprising five classes: one benign class, two DDoS attack types (DDoS-ICMP\_Flood and DDoS-UDP\_Flood), and two Mirai attack types (Mirai-UDPPlain and Mirai-Greeth\_Flood). The rationale for this selection is to retain only collective anomalies (DDoS and Mirai), as this study focuses on detecting collective anomalies. To reflect a realistic scenario, the new dataset is unbalanced; following the work in \cite{yu2015glad}, we use $20\%$ anomalous data and $80\%$ normal data.

A subset of the ToN-IoT network dataset is selected, including benign traffic and attack types such as DoS, DDoS, and scanning, as these represent collective anomalies.

Since our proposed method requires a graph as input, we employ the k-NN algorithm to reconstruct graphs from the data points, where each network packet is considered as a node in the generated graph. To ensure sparsity, a relatively low value of $k$ is used, resulting in a graph that is sparse by design. Sparse graphs are particularly advantageous for training, as they reduce computational complexity and memory requirements. Furthermore, sparsity enhances scalability, enabling the method to handle larger graphs efficiently. In this study, we generate three graphs corresponding to different subset sizes from the CICIoT2023 dataset, denoted as $G_{1}$, $G_{2}$, and $G_{3}$, to evaluate the scalability of our method. Using the ToN-IoT network dataset, four graphs $G'_{1}$, $G'_{2}$, $G'_{3}$, and $G'_{4}$ with different anomaly ratios were constructed to assess the robustness of the proposed method. Detailed information regarding the number of nodes and edges in each graph is reported in Table \ref{tab:graph_statistics}.

\begin{table}[ht!]
\caption{Statistics about the generated graphs.\label{tab:graph_statistics}}
\centering
\begin{tabular}{|c|c|c|c|c|}
\hline
\textbf{Graph} & \textbf{\# Nodes} & \textbf{\# Edges} & \textbf{\# Normal} & \textbf{\# Anomalous} \\
\hline
$G_{1}$ & 12,500 & 31,250 & 10,000 & 2,500 \\
\hline
$G_{2}$ & 125,000 & 312,500 & 100,000 & 25,000 \\
\hline
$G_{3}$ & 1,250,000 & 368,735 & 1,000,000 & 250,000 \\
\hline
$G'_{1}$ & 12,000 & 60,000 & 96,000 & 2,400 \\
\hline
$G'_{2}$ & 16,000 & 80,000    & 13,600 & 2,400 \\
\hline
$G'_{3}$ & 24,000 & 120,000 & 21,600 & 2,400 \\
\hline
$G'_{4}$ & 46,041 & 230,205 & 43,641 & 2,400 \\
\hline
\end{tabular}
\end{table}

\subsection{Experiment Reproducibility}

For reproducibility, in this part, we provide the parameters of our approach. Graphs were generated using the k-NN algorithm implemented with the FAISS library \cite{douze2024faiss}, which efficiently scales k-NN to large datasets. The VGAE was not fine-tuned; all parameters were kept constant across different input graphs. Specifically, the embedding size was set to 32 for the CICIoT2023 dataset and 64 for ToN-IoT network dataset, dropout to 0.2, learning rate to 0.001, and each model was trained for 200 epochs. For the first graph $G$ generated from raw data, we set $k = 5$ with the Euclidean distance metric. For the second graph $G'$ generated from embeddings, we used a higher value for $k$ with cosine distance metric, reflecting the higher similarity of the embeddings compared to the raw data. The confidence level was set to $\beta = 0.95$. Parameters not explicitly mentioned here were varied in the experiments, and their respective values are reported in the results.

\subsection{Comparison Methods}

To ensure a fair comparison, we conduct three types of evaluation. First, we compare our proposed method against traditional anomaly detection techniques, including LOF, IF, KNN, and OCSVM, using the CICIoT23 and ToN-IoT network datasets. Second, we evaluate our method against clustering-based CAD approaches on the CICIoT23 dataset, and further assess its scalability. More specifically, we compare with five representative clustering-based CAD methods: CADET, HCADET, MCADET, HDM-GAD, and CCAD. Third, we perform a robustness study by evaluating our method under varying anomaly ratios and comparing it with CAD methods on the ToN-IoT network dataset.

\subsection{Evaluation Metrics}

We evaluate our proposed method on standard DL evaluation metrics such as accuracy, recall, f1-score, area under the ROC curve (auc), and the area under the precision–recall curve (PR). In addition to these metrics, we used purity, which is defined in Eq. \ref{EqPurity} as the extent to which each cluster contains nodes from a single class, in this case, either normal or collective anomaly.

\begin{equation} \label{EqPurity}
Purity = \frac{1}{N} \sum_{k} \max_{j} |c_k \cap t_j|
\end{equation}

where $N$ is the total number of nodes, $c_k$ is the set of nodes in cluster $k$, and $t_j$ is the set of nodes belonging to class $j$. The value of purity ranges from 0 to 1, with higher values indicating that clusters are more homogeneous with respect to the ground truth.

\section{Performance Evaluation} \label{sec:pe}

In this section, we present a series of experiments to evaluate the effectiveness of our proposed method, UGCAD, and to compare it against state-of-the-art approaches. Specifically, we address the following research questions:

\begin{itemize}
    \item \textbf{RQ1:} How does UGCAD perform compared to traditional anomaly detection methods?
    \item \textbf{RQ2:} How effective is the EDC module in clustering network traffic data?
    \item \textbf{RQ3:} How do hyperparameters influence the performance of the CRAD module?
    \item \textbf{RQ4:} How does UGCAD perform in comparison to state-of-the-art CAD methods?
    \item \textbf{RQ5:} How does UGCAD perform across diverse datasets, and how robust is it to variations in anomaly ratios?
\end{itemize}

\subsection{Performance Comparison with Classical Anomaly Detection Methods(RQ1) }

\begin{table}[ht!]
\centering
\tiny 
\caption{The Evaluation Results of UGCAD versus Traditional Anomaly Detection Methods.}
\label{tab:comparison_trad}
\begin{tabular}{|l|l|c|c|c|c|c|}
\hline
\textbf{Dataset} & \textbf{Metric} & \textbf{KNN} & \textbf{OCSVM} & \textbf{LOF} & \textbf{IF}  &  \textbf{Ours}  \\
\hline
\multirow{2}{*}{CICIoT2023} 
& AUC  & 0.411 & 0.503  & 0.754 & 0.481 & \textbf{\textasciitilde 1}  \\

& PR  & 0.198 & 0.198  & 0.560 & 0.204 & \textbf{\textasciitilde 1}  \\
\hline

\multirow{2}{*}{TON-IoT} 
& AUC  & 0.319 & 0.495  & 0.583 & 0.535 &\textbf{0.61} \\
& PR  & 0.1474 & 0.212 & 0.267 & 0.224 &\textbf{0.557} \\

\hline

\end{tabular}
\end{table}

A comparative analysis of UGCAD and traditional anomaly detection methods, including KNN, OCSVM, LOF, and IF, is presented in Table \ref{tab:comparison_trad}, across two datasets and more specifically on graphs $G_{1}$ and $G'{1}$ using PR and AUC as evaluation metrics. Since this part of the evaluation is conducted in a threshold-independent manner, metrics that require a fixed decision threshold, such as accuracy, are not considered.


On the CICIoT2023 dataset, the proposed method achieves near-perfect performance, with both AUC and PR values approaching 1. This indicates an almost complete separability between normal and anomalous instances, suggesting that the proposed approach is highly effective at capturing the underlying data distribution in this setting. In contrast, all baseline methods exhibit substantially inferior performance. Among them, LOF attains the highest scores (AUC = 0.754, PR = 0.560), yet remains significantly below the proposed method. The remaining approaches yield AUC values close to random performance and consistently low PR scores, highlighting their limited capability in modeling the structure of this dataset.

On the more challenging ToN-IoT network dataset, a noticeable degradation in performance is observed across all methods, reflecting the increased difficulty of the dataset, where normal and anomalous samples exhibit greater overlap. The proposed method achieves an AUC of 0.61 and a PR of 0.557. Although lower than the results obtained on CICIoT2023, these scores remain the highest among all evaluated methods. LOF again constitutes the strongest baseline (AUC = 0.583, PR = 0.267), but a substantial gap persists, particularly in terms of PR. The generally lower performance across all methods suggests that ToN-IoT represents a more complex and realistic anomaly detection scenario.

Overall, these results demonstrate that the proposed method consistently outperforms traditional anomaly detection techniques across both datasets, which answers RQ1. Furthermore, the observed performance drop on ToN-IoT dataset underscores the impact of dataset characteristics, particularly class overlap, on detection performance. Despite these challenges, the proposed approach maintains superior discriminative capability, indicating its robustness and effectiveness in comparison to existing methods.

\subsection{Performance Evaluation of the EDC module (RQ2)}

After evaluating the proposed method UGCAD against traditional approaches, we proceed to assess its individual components on the CICIoT2023 dataset. We begin by evaluating the EDC module.

The first module, EDC, is responsible for generating node embeddings and clustering them into meaningful groups. The underlying assumption is that node embeddings enhance cluster quality and facilitate the separation of normal data from collective anomalies. Ideally, each cluster should predominantly contain nodes of the same class, resulting in distinct clusters that represent either collective anomalies or normal data. To assess the quality of the clustering, we employ the purity metric, which quantifies the extent to which a cluster is dominated by a single class, whether anomalous or normal. Table~\ref{tab:purity_results} presents the purity scores for the three graphs. As observed, the EDC module effectively separates nodes into clusters with a clear majority class. This strong separation is crucial for the subsequent module, which performs anomaly detection, to operate effectively. These results address RQ2 by demonstrating the effectiveness of the EDC module in clustering network traffic data. 

\begin{table}[ht!]
\caption{Performance of the EDC Module w.r.t Cluster Purity.
\label{tab:purity_results}}
\centering
\begin{tabular}{|c|c|c|c|}
\hline
\textbf{Dataset} & $G_{1}$ & $G_{2}$ & $G_{3}$\\
\hline
\textbf{Purity} & 0.996 & 0.997  &  0.998  \\
\hline
\end{tabular}
\end{table}

\subsection{Parameter Analysis (RQ3)}

\begin{center}

\includegraphics[width=2.5in]{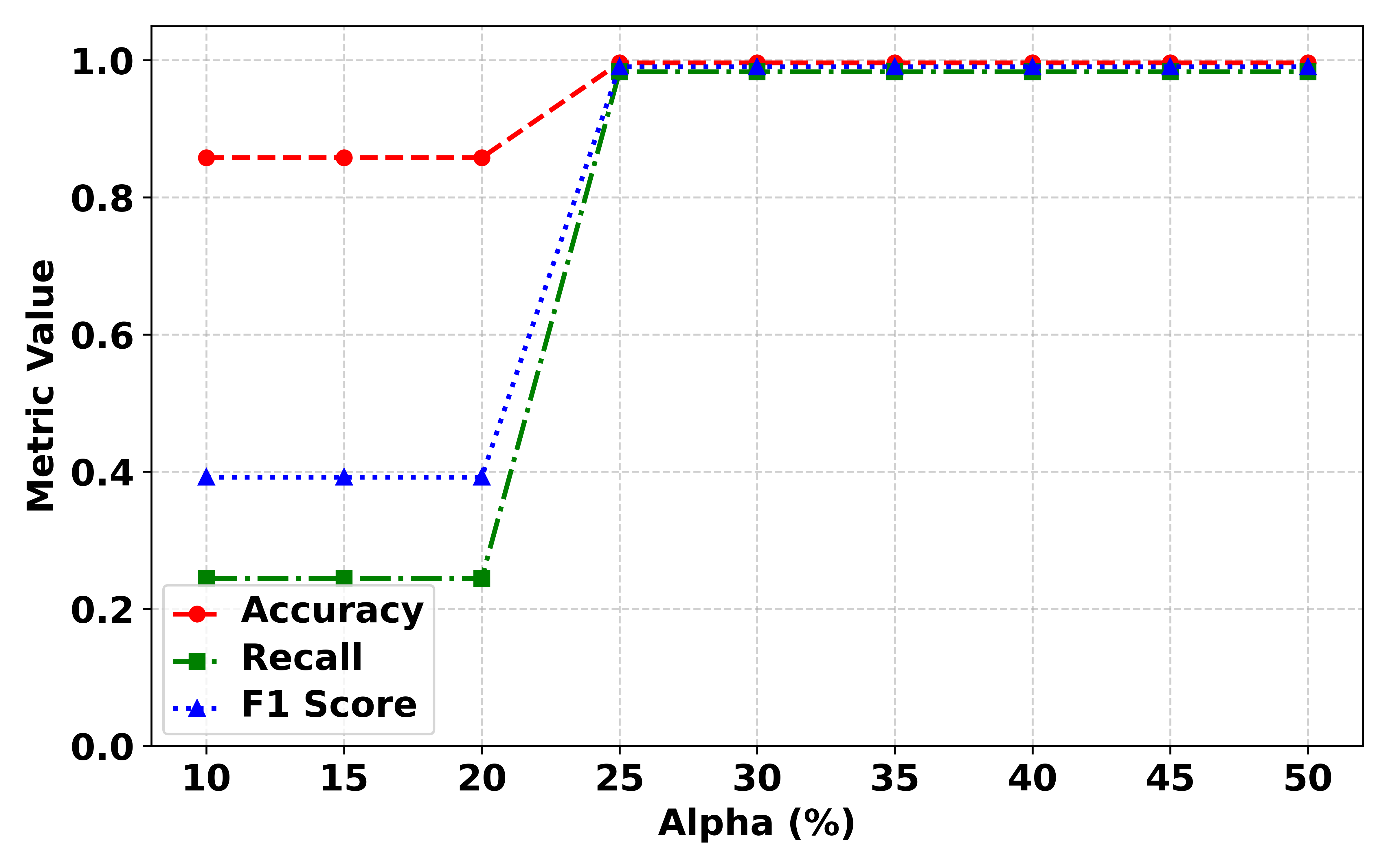}

(a)

\vspace{1ex}

\includegraphics[width=2.5in]{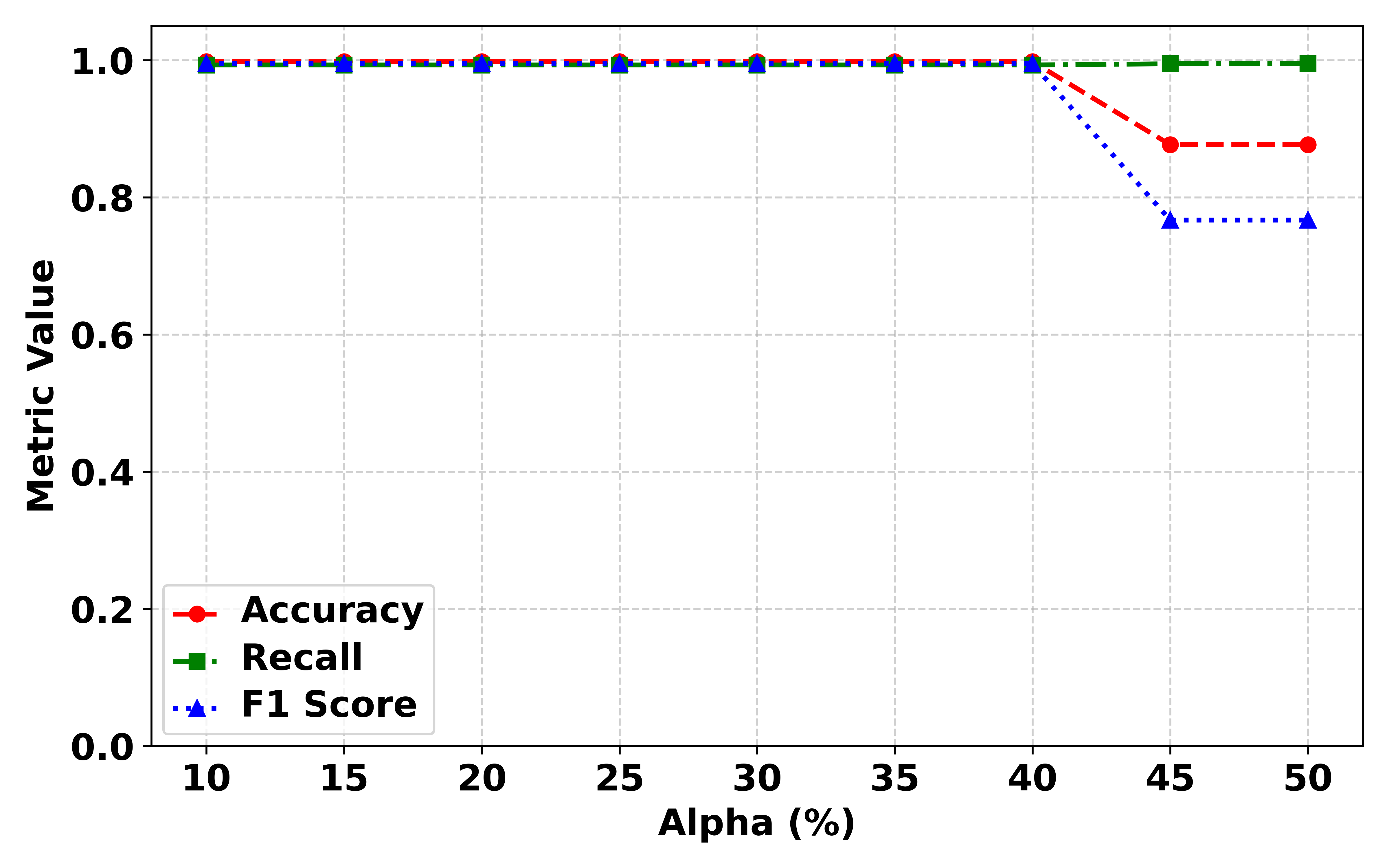}

(b)

\vspace{1ex}

\includegraphics[width=2.5in]{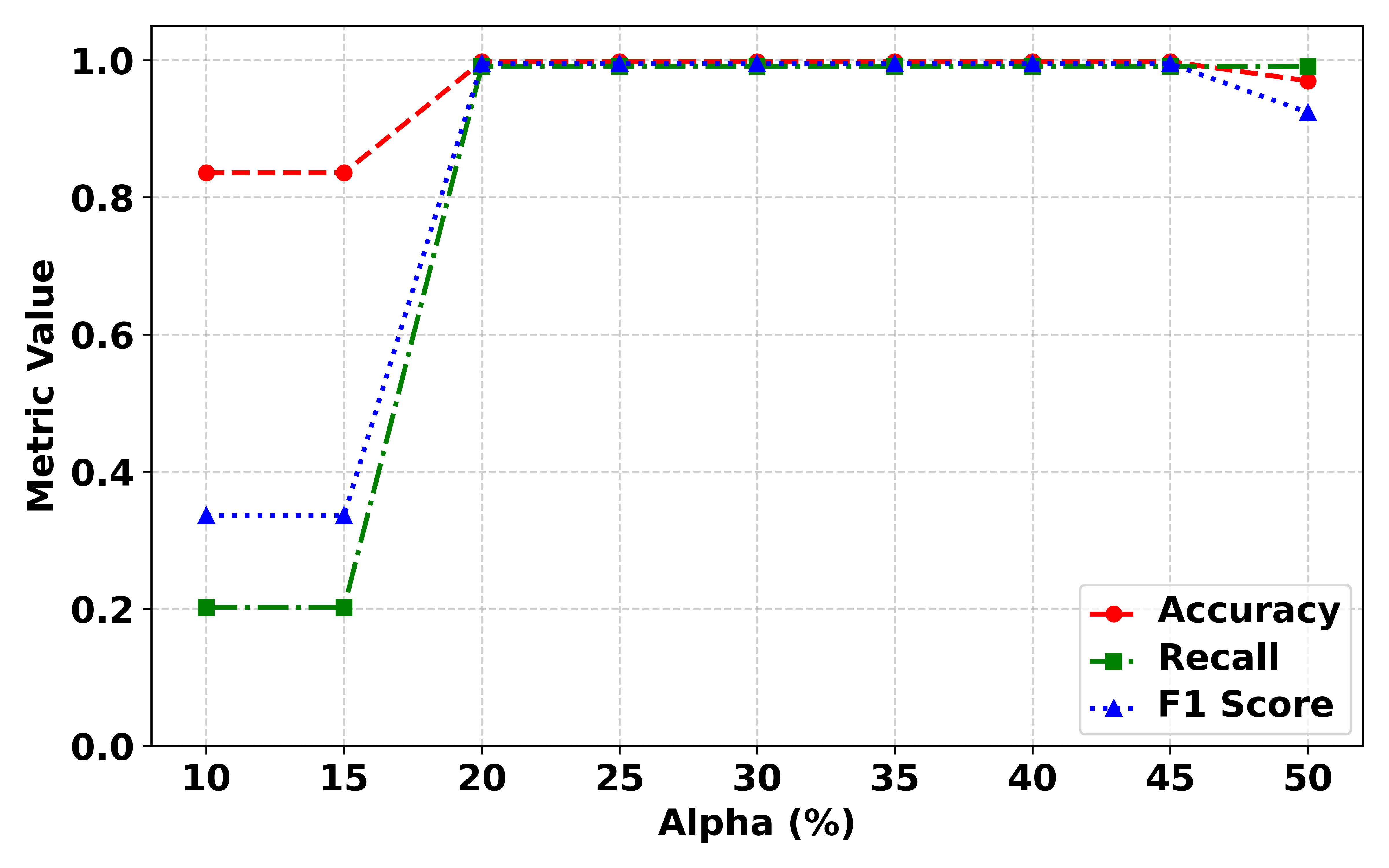}

(c)

\captionof{figure}{Experimental results for parameter analysis. (a), (b), and (c) show the impact of parameter $\alpha$ on different metrics for graphs $G_{1}$, $G_{2}$, and $G_{3}$, respectively.}

\label{fig:alpha_impact}

\end{center}

In this part, the parameter $\alpha$ is analyzed, which represents the percentile of the cluster score in $\mathcal{A}$ used to refine normal clusters in $\mathcal{R}$, producing $\mathcal{R'}$. Adjusting $\alpha$ aims to select the score threshold $T_{\alpha}$ at which $\mathcal{R'}$ most accurately reflects the anomaly scores of normal data. Typically, $\alpha$ corresponds to a low percentile, since the objective is not to refine the anomalous clusters in $\mathcal{R}$ to the extent that only the very lowest anomaly scores remain in $R'$. Thus, carefully balancing $\alpha$ is crucial to ensure $\mathcal{R'}$ effectively captures the distinction between normal and anomalous nodes.

For the experiments, we vary $\alpha$ from 10\% to 50\%, as values outside this range are inconsistent with the intended purpose of $\alpha$. Fig.~\ref{fig:alpha_impact} illustrates the effect of varying $\alpha$ on accuracy, recall, F1-score for the graphs $G_{1}$, $G_{2}$, and $G_{3}$. The results indicate that for $G_{1}$ and $G_{3}$, the metrics initially start at relatively low values but quickly stabilize at high values exceeding 99\%. In contrast, for $G_{2}$, the metrics begin high at approximately 99\% and decline when $\alpha$ exceeds 40\%. Based on these observations, we conclude that $\alpha$ values of 25\% to 40\% achieve the best performance across all graphs and this answers RQ3. We adopt this range for the remainder of the experiments.

\subsection{Performance Comparison (RQ4)}

\begin{figure}[t]
\centering

\begin{minipage}[]{\linewidth}
\centering
\includegraphics[width=2.5in]{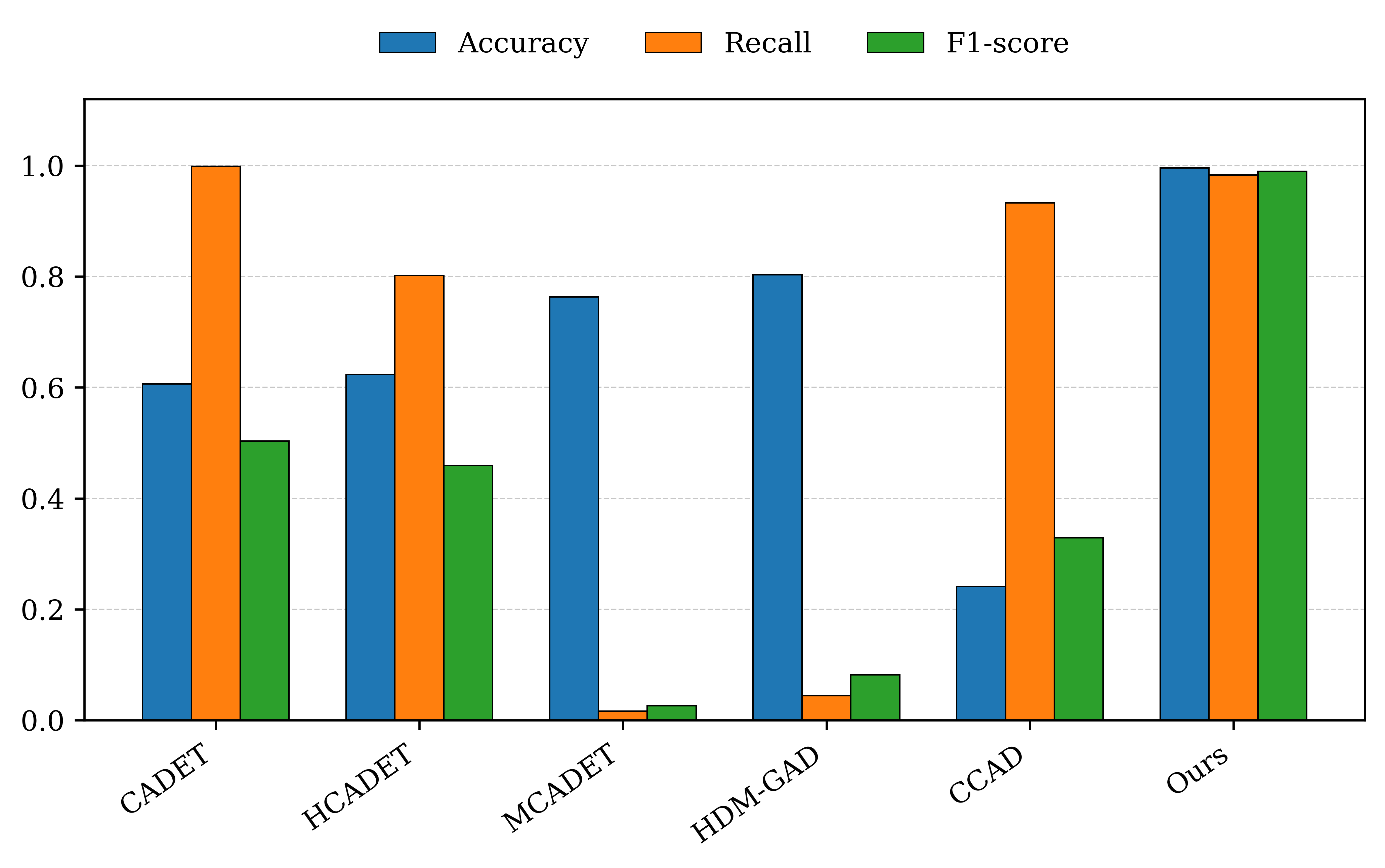}
\subcaption*{(a)}
\label{G1_res}
\end{minipage}

\vspace{1ex}

\begin{minipage}{\linewidth}
\centering
\includegraphics[width=2.5in]{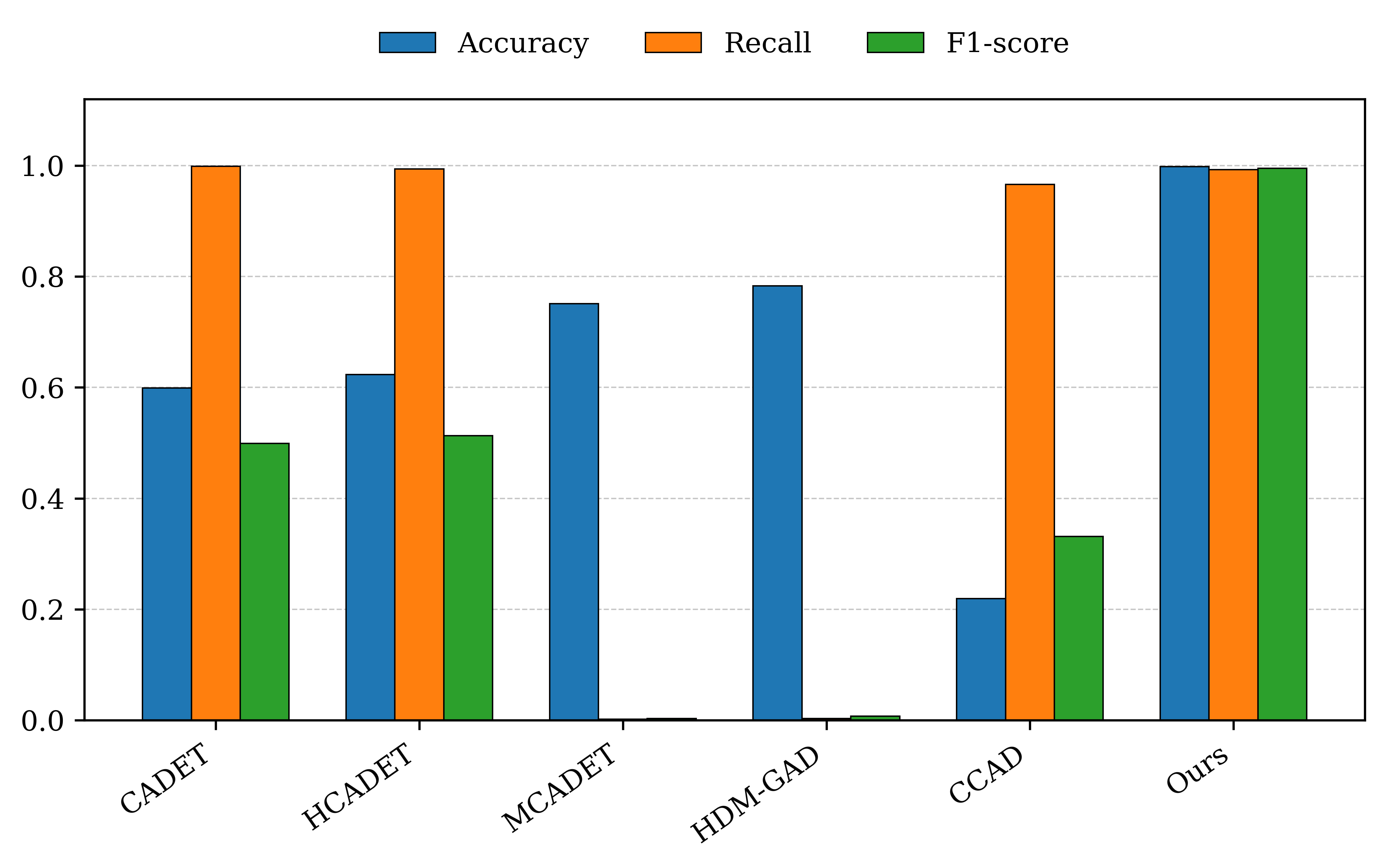}
\subcaption*{(b)}
\label{G2_res}
\end{minipage}

\vspace{1ex}

\begin{minipage}{\linewidth}
\centering
\includegraphics[width=2.5in]{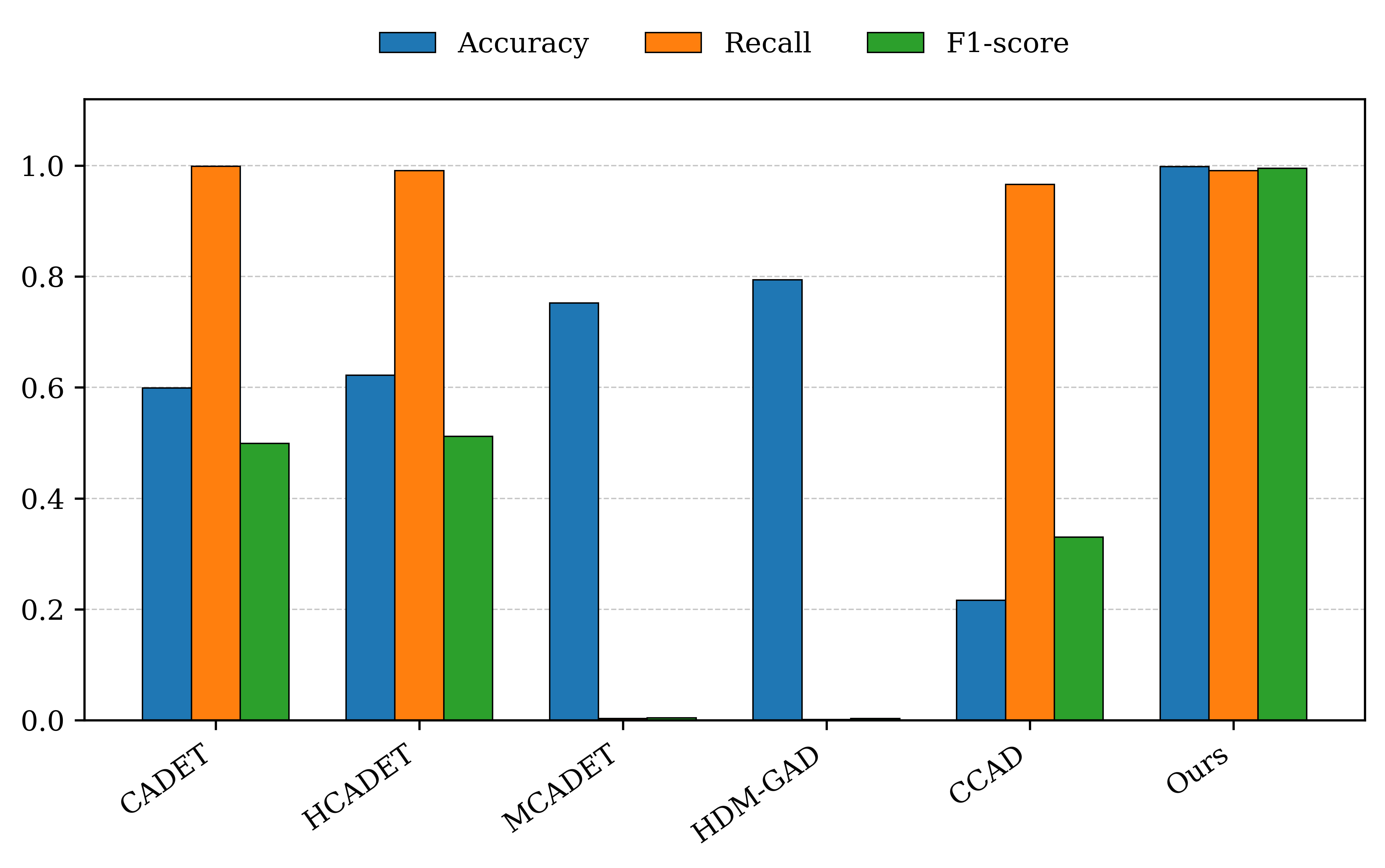}
\subcaption*{(c)}
\label{G3_res}
\end{minipage}

\caption{Performance comparison of different methods on the three graphs $G_{1}$, $G_{2}$, and $G_{3}$, respectively}
\label{fig:comparison}

\end{figure}

This section evaluates the proposed method UGCAD, against other clustering-based CAD approaches to address RQ4. Fig.~\ref{fig:comparison} reports the comparison results in terms of accuracy, F1-score, and recall across the three graphs. 
From Fig.~\ref{fig:comparison}, we can draw the following conclusions:  

1) UGCAD achieves the best accuracy and F1-score, all exceeding 99\%, and ranks second in recall (after CADET) on the $G_{1}$ and $G_{3}$ graphs. This directly answers RQ4, confirming the effectiveness of UGCAD in detecting collective anomalies while maintaining a low false alarm rate.  

2) CADET consistently achieves the best recall across all graphs, indicating strong anomaly detection capability. However, its F1-score remains average despite a decent accuracy , suggesting a relatively high false alarm rate.  

3) HCADET shows comparable performance, with solid accuracy. On the $G_{2}$ dataset, it achieves the highest recall (over 99\%) after CADET, but again, its average F1-score indicates a high false alarm rate. 

4) MCADET and HDM-GAD deliver reasonable accuracy (above 0.75). However, their recall and F1-scores are consistently low across all graphs, pointing to both a high false positive rate and a low true positive rate, meaning the anomaly detection task was not performed effectively.  

5) CCAD shows low accuracy and F1-score but very high recall on all graphs. This happens because the sliding window method flags most windows as anomalous, detects many collective anomalies (high recall), but also misclassifies many normal samples (low accuracy).

\hspace{1cm}
Overall, while most methods attain satisfactory accuracy, this is partly due to the predominance of normal nodes in the graphs. Recall and F1-score provide more reliable indicators of anomaly detection quality. UGCAD not only achieves the highest accuracy but also demonstrates superior recall and F1-scores (exceeding 0.99), highlighting its scalability and effectiveness for the CAD task.  

\subsection{Generalization and Robustness Analysis}

\begin{center}

\includegraphics[width=2.5in] {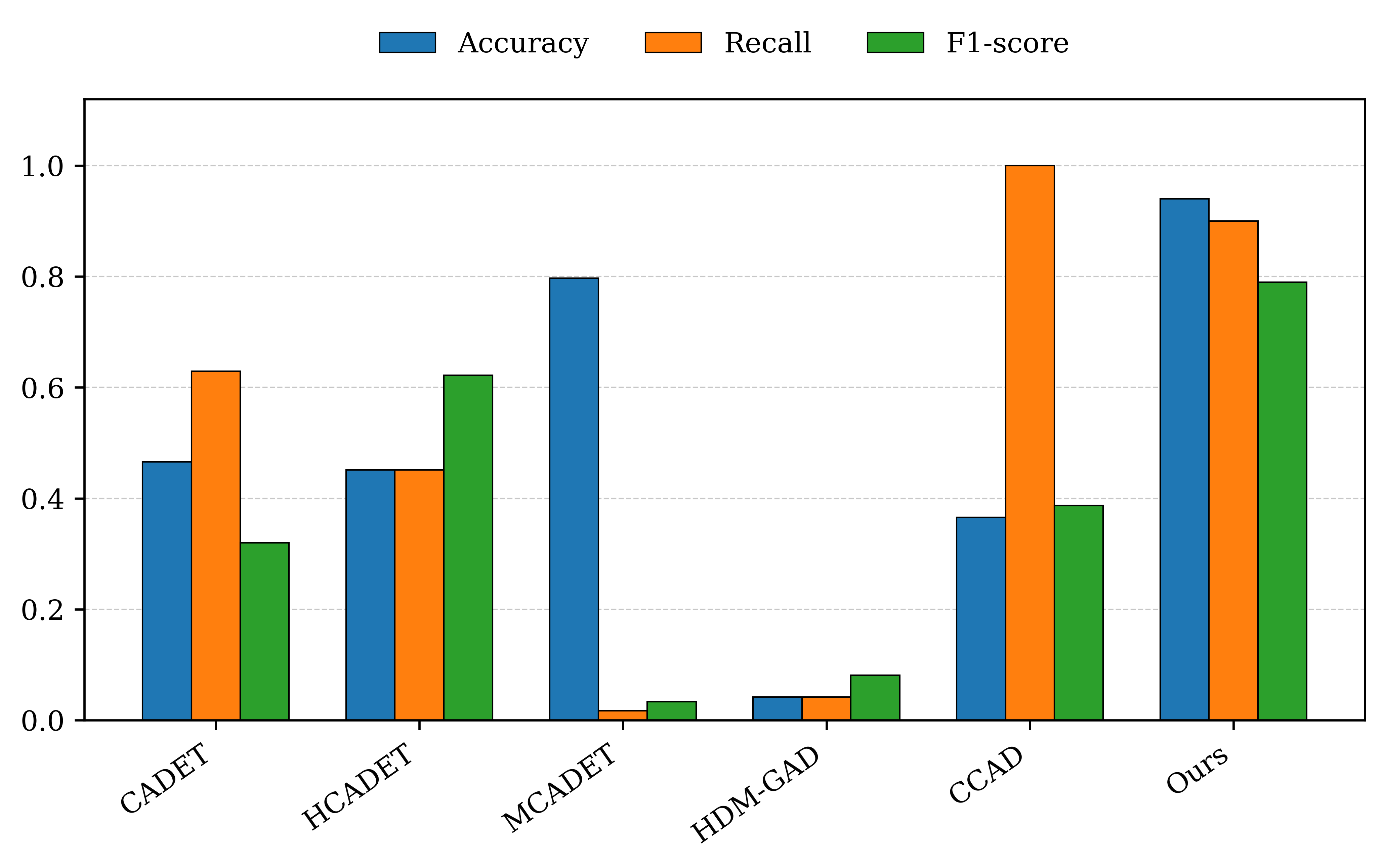}
\\
(a)

\includegraphics[width=2.5in]{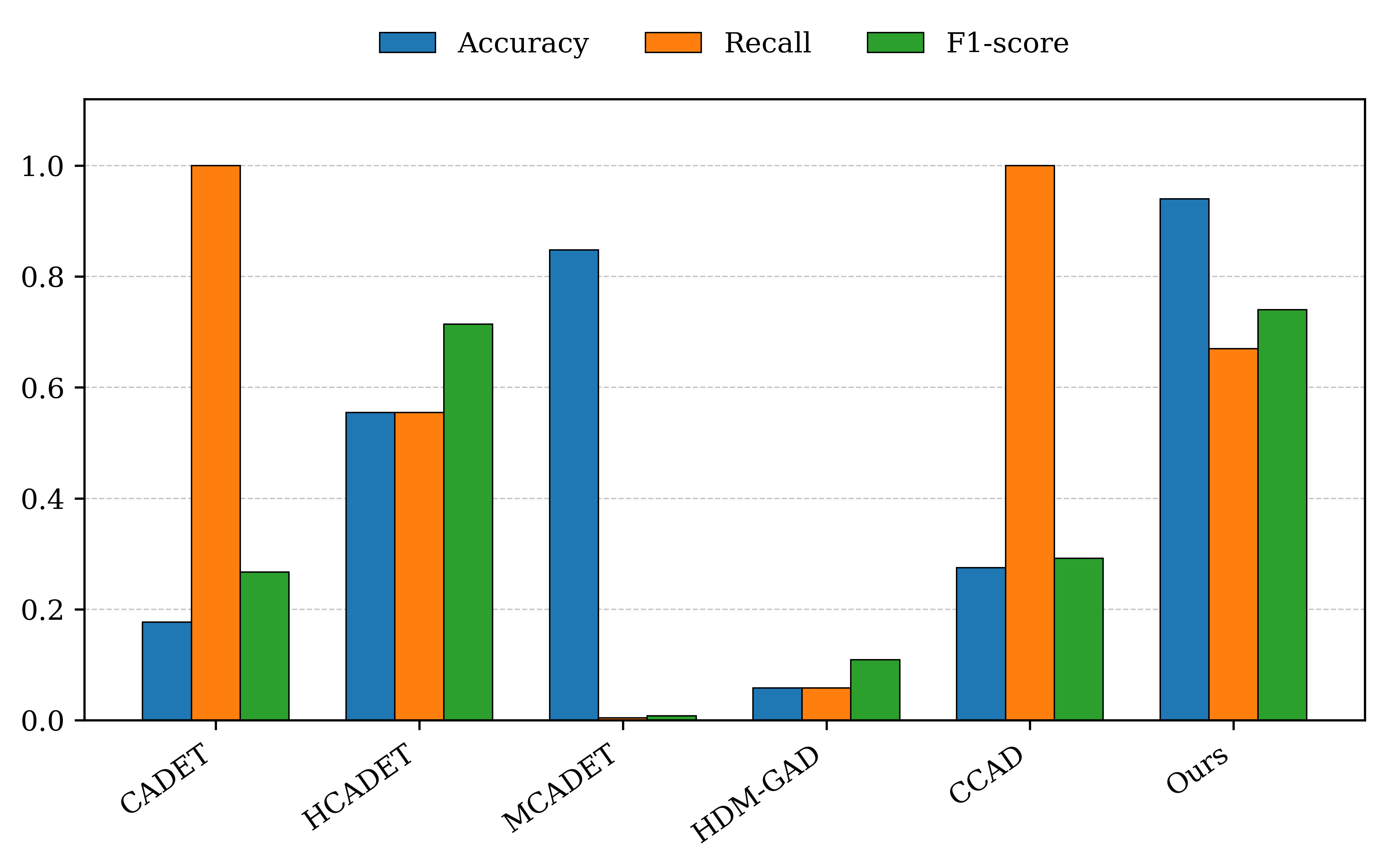}

(b)


\includegraphics[width=2.5in]{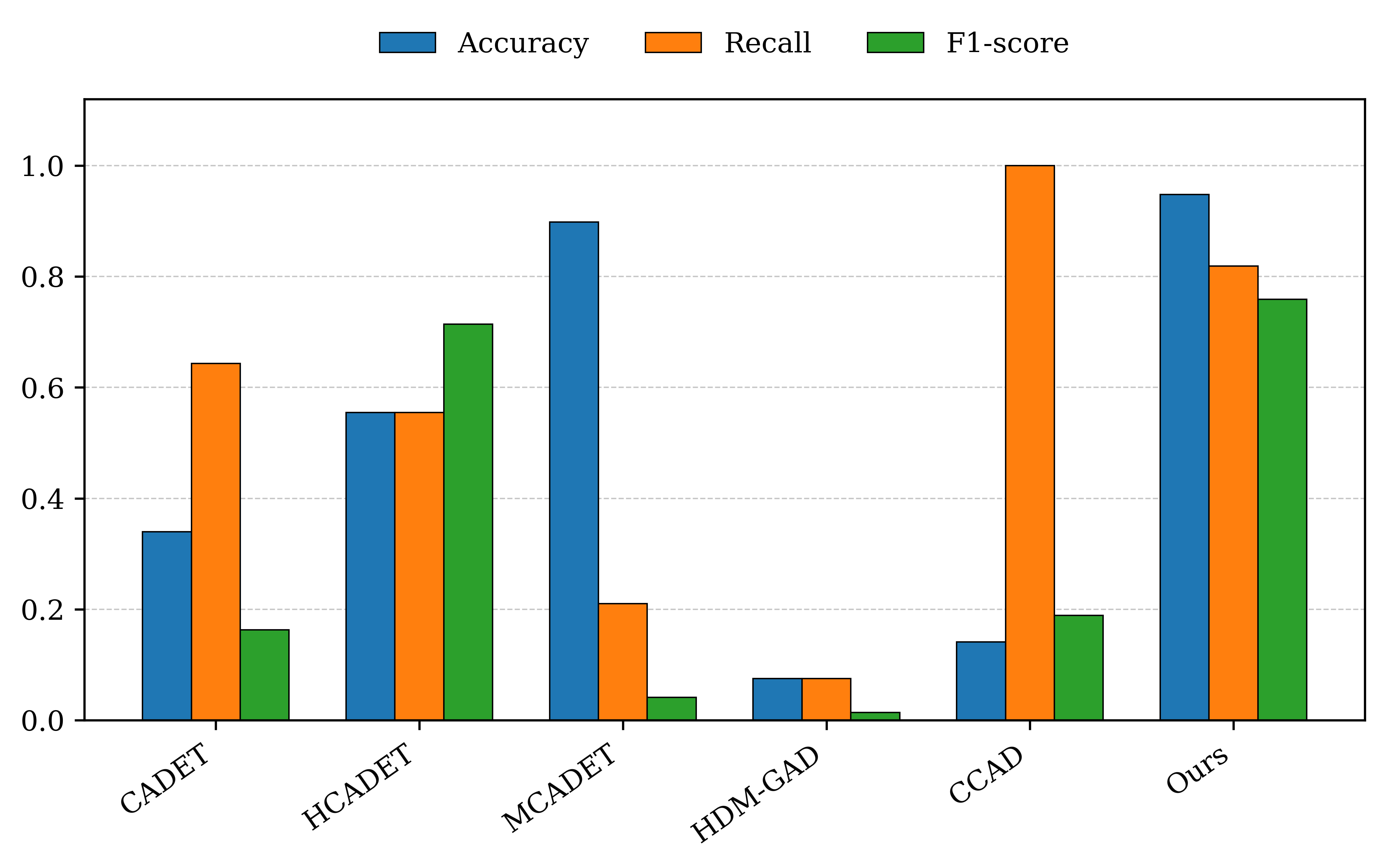}

(c)

\vspace{1ex}

\includegraphics[width=2.5in]{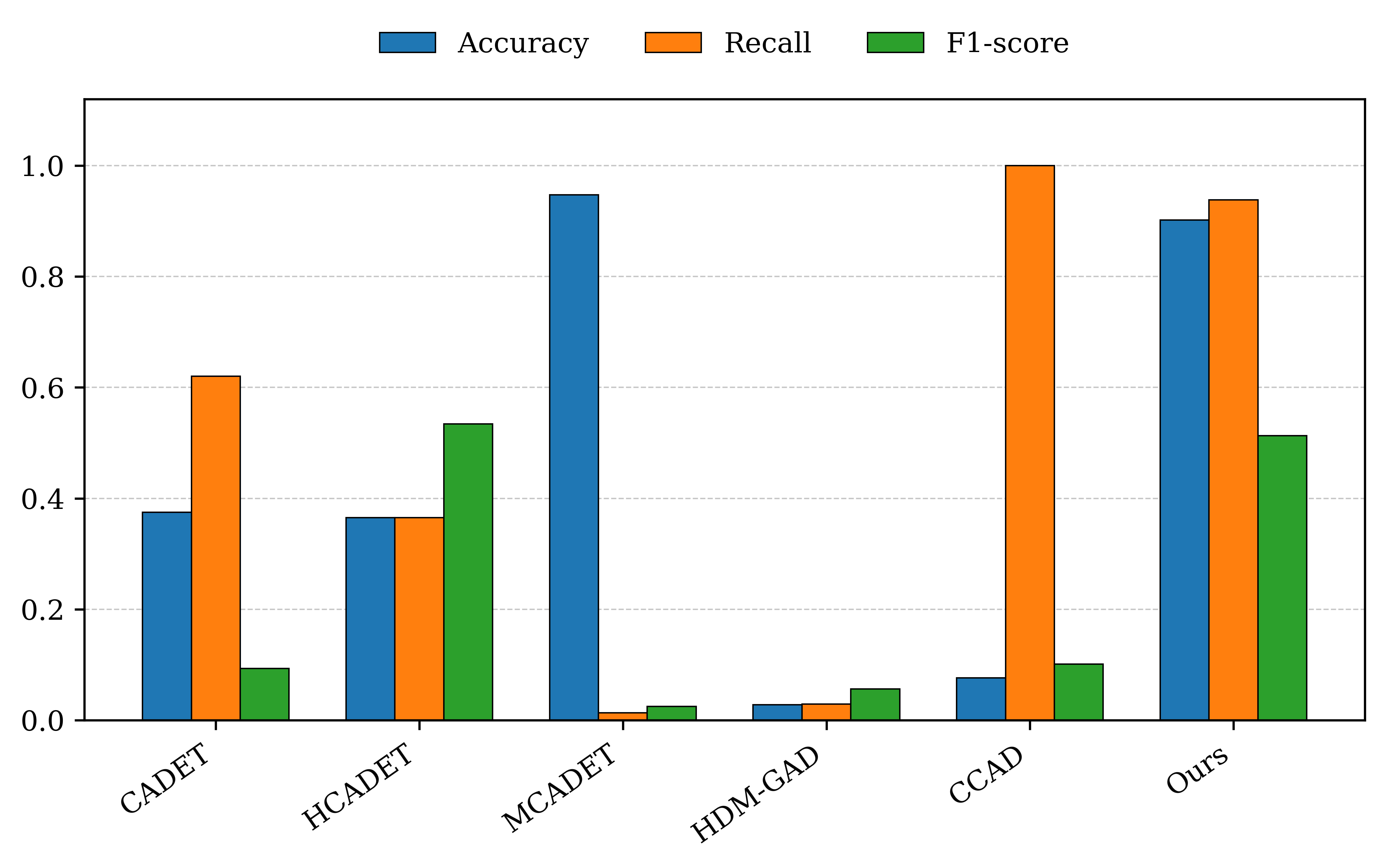}

(d)

\captionof{figure}{Performance comparison of CAD methods on the graphs $G'_{1}$, $G'_{2}$, $G'_{3}$, and $G'_{4}$, respectively.}

\label{fig:comparison_ton}

\end{center}

After evaluating the performance of UGCAD on the CICIoT2023 dataset with an anomaly ratio of 20\% and comparing it against existing CAD methods, we extend the evaluation to the ToN-IoT network dataset under varying anomaly ratios. To reflect more realistic scenarios, the proportion of anomalies is varied from 20\% to 5\%. Specifically, the graphs ($G'_{1}$), ($G'_{2}$), ($G'_{3}$), and ($G'_{4}$) correspond to anomaly ratios of 20\%, 15\%, 10\%, and 5\%, respectively. The corresponding results are presented in Fig. \ref{fig:comparison_ton}.

At higher anomaly ratios ($G'_{1}$ , 20\%), UGCAD significantly outperforms all baseline methods in terms of accuracy (0.94) and F1-score (0.79), while maintaining a high recall (0.9). Among the baselines, MCADET achieves relatively high accuracy (0.797), but its extremely low recall (0.017) indicates poor anomaly detection capability. In contrast, CCAD achieves perfect recall (1.0), but suffers from low accuracy and F1-score, suggesting a bias toward predicting anomalies.

As the anomaly ratio decreases to 15\% ($G'_{2}$ ), similar trends persist. UGCAD maintains superior overall performance, achieving the highest accuracy (0.94) and F1-score (0.74), while preserving a balanced recall (0.67). MCADET again shows high accuracy (0.848) but near-zero recall, confirming its inability to detect rare anomalies. HCADET provides a more balanced performance among baselines, achieving the highest F1-score (0.714) in this setting, though still below UGCAD.

At lower anomaly ratios ($G'_{3}$, 10\% and $G'_{4}$, 5\%), the performance gap becomes more pronounced. UGCAD remains stable, achieving the highest accuracy and F1-score across both graphs, and maintaining strong recall (0.819 and 0.938, respectively). In contrast, most baseline methods degrade significantly. MCADET continues to report high accuracy (up to 0.947), but its recall drops to near zero, indicating that it largely fails to identify anomalies as they become rarer. HDM-GAD consistently exhibits poor performance across all metrics and graphs. CADET shows moderate recall but suffers from very low F1-scores, reflecting poor precision.

HCADET remains the most stable baseline, maintaining relatively consistent F1-scores across graphs; however, it does not match the overall performance of UGCAD. Results for CCAD degrade as the anomaly ratio decreases on both $G'_{3}$ and $G'_{4}$, suggesting a tendency toward high Recall at the expense of precision. 

In summary, the proposed method demonstrates superior robustness to varying anomaly ratios, maintaining a strong balance between accuracy, recall, and F1-score. In contrast, baseline methods either overfit to majority classes or over-predict anomalies, leading to unstable performance as the anomaly ratio decreases.

\subsection{Results Analysis}
\begin{figure}[!t]
\centering
\includegraphics[width=3in, height=0.2\textheight]{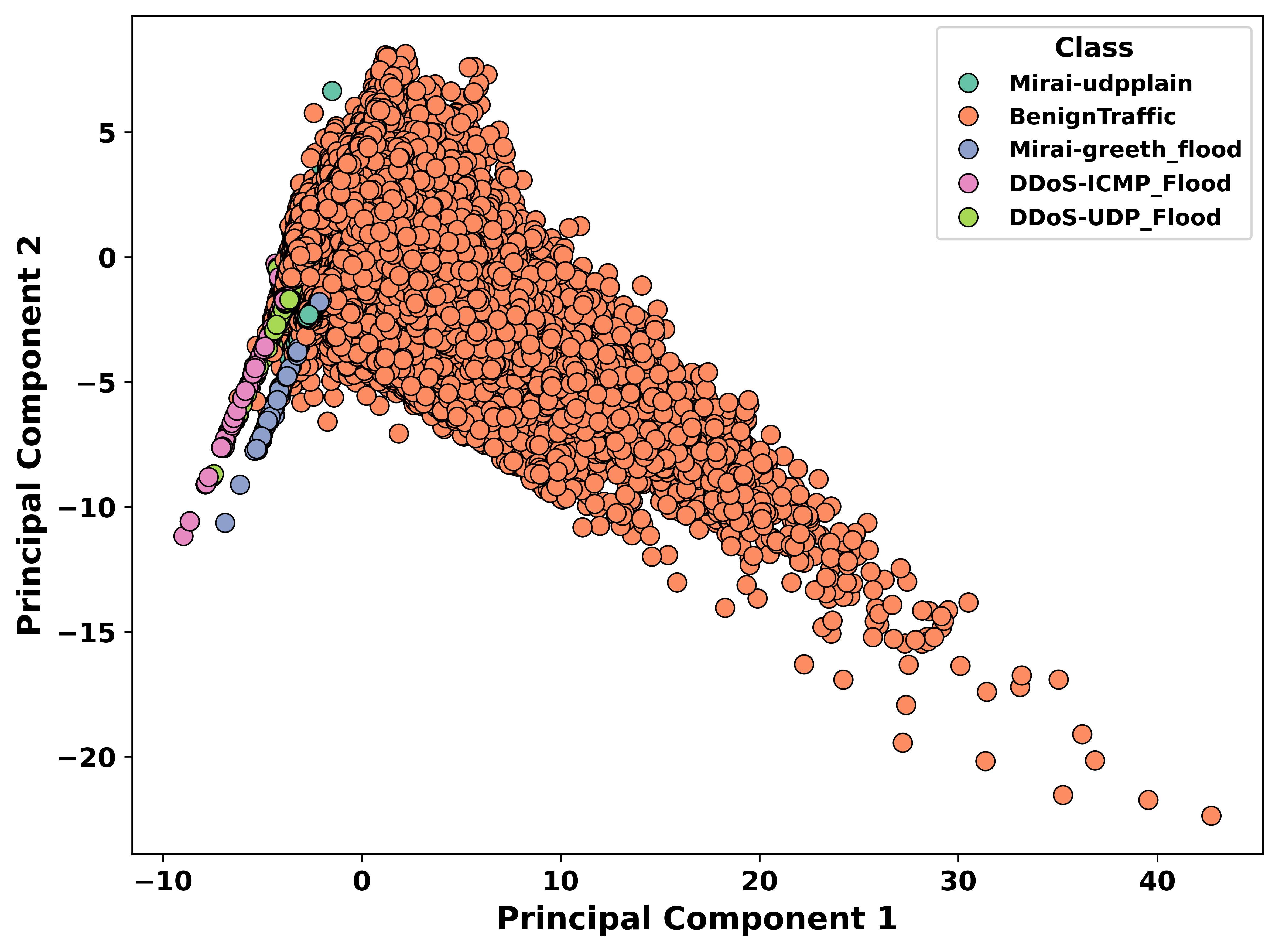}
\caption{Visualization on the $G_{3}$ dataset using PCA. }
\label{fig:vis-pca}
\end{figure}
Collective anomalies are not merely aggregations of point anomalies that can be easily separated from normal data. Instead, they exhibit group-level properties that make them appear similar to normal samples. Fig.~\ref{fig:vis-pca} presents the PCA projection of the $G_{3}$ graph, where benign and attack samples show some separation. However, the overlap among anomalous and normal samples highlights the difficulty of distinguishing collective anomalies from normal data. This challenge is particularly evident in clustering-based CAD, which relies heavily on cluster quality. Poorly formed clusters, lacking purity (i.e., containing samples from multiple classes), degrade detection performance and lead to weak evaluation metrics. UGCAD addresses this issue by leveraging node embeddings rather than raw features, enabling more robust clusters. In contrast, methods that rely purely on statistical properties such as density or variance struggle with CAD, as normal data may exhibit similar properties. This often results in high false alarm rates, as seen in approaches like HDM-GAD that relies on cluster density. 

In particular, compared to traditional methods such as KNN, OCSVM, LOF, and IF, UGCAD consistently achieves superior performance on both the CICIoT2023 and ToN-IoT datasets. UGCAD obtains significantly higher AUC and PR values, demonstrating a much stronger capability in identifying collective anomalies. In contrast, traditional approaches exhibit limited effectiveness, especially in terms of PR, as they are primarily designed for point anomaly detection and fail to capture the structural and relational characteristics inherent in graph-based collective anomalies.

When compared to existing CAD approaches, several limitations become evident. MCADET attains relatively high accuracy (around 0.75 on CICIoT2023 and up to 0.85–0.95 on ToN-IoT as the anomaly ratio decreases), yet its recall and F1-score remain extremely low across all experiments. This indicates a strong bias toward the majority class, where the method predominantly predicts normal samples and fails to detect anomalies. This behaviour stems from its multi-stage clustering strategy, where low-variance clusters, more likely representing normal behaviour in imbalanced settings, are incorrectly selected as anomalies, leading to deteriorating detection performance despite increasing accuracy.

CADET, on the other hand, achieves near-perfect recall on CICIoT2023, but this comes at the cost of moderate accuracy and F1-score, reflecting a high number of false positives due to its strong bias toward anomaly prediction. Its performance remains largely invariant to graph size. However, on ToN-IoT, its performance degrades significantly as the anomaly ratio decreases, with unstable recall, declining F1-score, and relatively low accuracy, highlighting its sensitivity to class imbalance. This is mainly due to its simplistic decision rule of selecting anomalous clusters based on size, which leads to overestimation of anomalies and reduced discrimination when anomalies are sparse.

HCADET demonstrates more stable behaviour. On CICIoT2023, recall improves with increasing graph size (from approximately 0.80 to nearly 0.99), although accuracy and F1-score remain moderate, indicating a continued bias toward anomaly detection. On ToN-IoT, the method maintains relatively stable performance across varying anomaly ratios, with accuracy around 0.55 and moderate F1-scores. This improved robustness is attributed to its use of Hurst-based statistical ranking of clusters. However, relying on global statistical properties limits its ability to sharply distinguish anomalies, resulting in moderate yet consistent performance rather than strong detection capability.

HDM-GAD exhibits extremely poor performance across both datasets. On CICIoT2023, all metrics (accuracy, recall, and F1-score) remain close to zero, indicating a complete failure to detect anomalies. Similarly, on ToN-IoT, performance remains consistently low across all anomaly ratios, with no noticeable improvement even at higher anomaly levels. This is due to its reliance on assumptions such as dense micro-clusters and meaningful sequential patterns for anomalies, which do not hold in network intrusion data that is typically sparse and heterogeneous. Consequently, its multi-stage framework fails to isolate anomalous groups and propagates incorrect detections.

CCAD achieves consistently very high recall (approximately 0.93–0.96) on CICIoT2023 across all graph sizes, but suffers from low accuracy and F1-score, indicating excessive false positives due to strong anomaly prediction bias. Its performance remains stable with respect to dataset size. On ToN-IoT, similar behaviour is observed at higher anomaly ratios (20\% and 15\%), where recall is nearly perfect but overall classification balance remains poor. On lower anomaly ratios (10\% and 5\%), CCAD continues to achieve high recall but other metrics degrade significally. This behaviour arises from its reliance on detecting clustering structure changes in sliding windows, which can also be triggered by normal variations in dynamic network traffic, leading to over-detection of anomalies.

Overall, existing CAD methods either suffer from strong bias toward the majority class (MCADET), excessive anomaly prediction (CADET and CCAD), limited discriminative power (HCADET), or fundamental design mismatches with the data characteristics (HDM-GAD). UGCAD consistently achieves superior and balanced performance. Its ability to effectively capture the structural properties of collective anomalies allows it to outperform both traditional and state-of-the-art CAD approaches across different datasets and varying anomaly conditions.

\section{Conclusion and future work} \label{sec: conc}
We proposed in this paper UGCAD, a framework for detecting collective anomalies in IoT network traffic. It consists of two main modules: EDC and CRAD. EDC employs a VGAE  to generate node-level embeddings, which are then used to enhance the Leiden algorithm for high-quality clustering of network traffic data. The CRAD module subsequently aggregates clusters identified as normal to serve as a reference and applies reconstruction error as an anomaly score to distinguish anomalous clusters from normal ones. The effectiveness of UGCAD was evaluated on the two datasets CICIoT2023 and ToN-IoT network dataset, and results demonstrated strong performance in both clustering quality and anomaly detection. Moreover, comparative experiments with several clustering-based CAD approaches and traditional methods confirm the superior performance of UGCAD. Despite its advantages, UGCAD has certain limitations.  Its reliance on clustering makes the method highly
dependent on the quality of embeddings, and because the framework is clustering-based, it cannot generalize naturally to unseen or streaming data, limiting its applicability for real-time anomaly detection on its own. Future research includes adapting this CAD strategy to naturally work on unseen data and detect collective anomalies in real-time. 


\begin{IEEEbiography}
[{\includegraphics[width=1in,height=1.25in,clip,keepaspectratio]{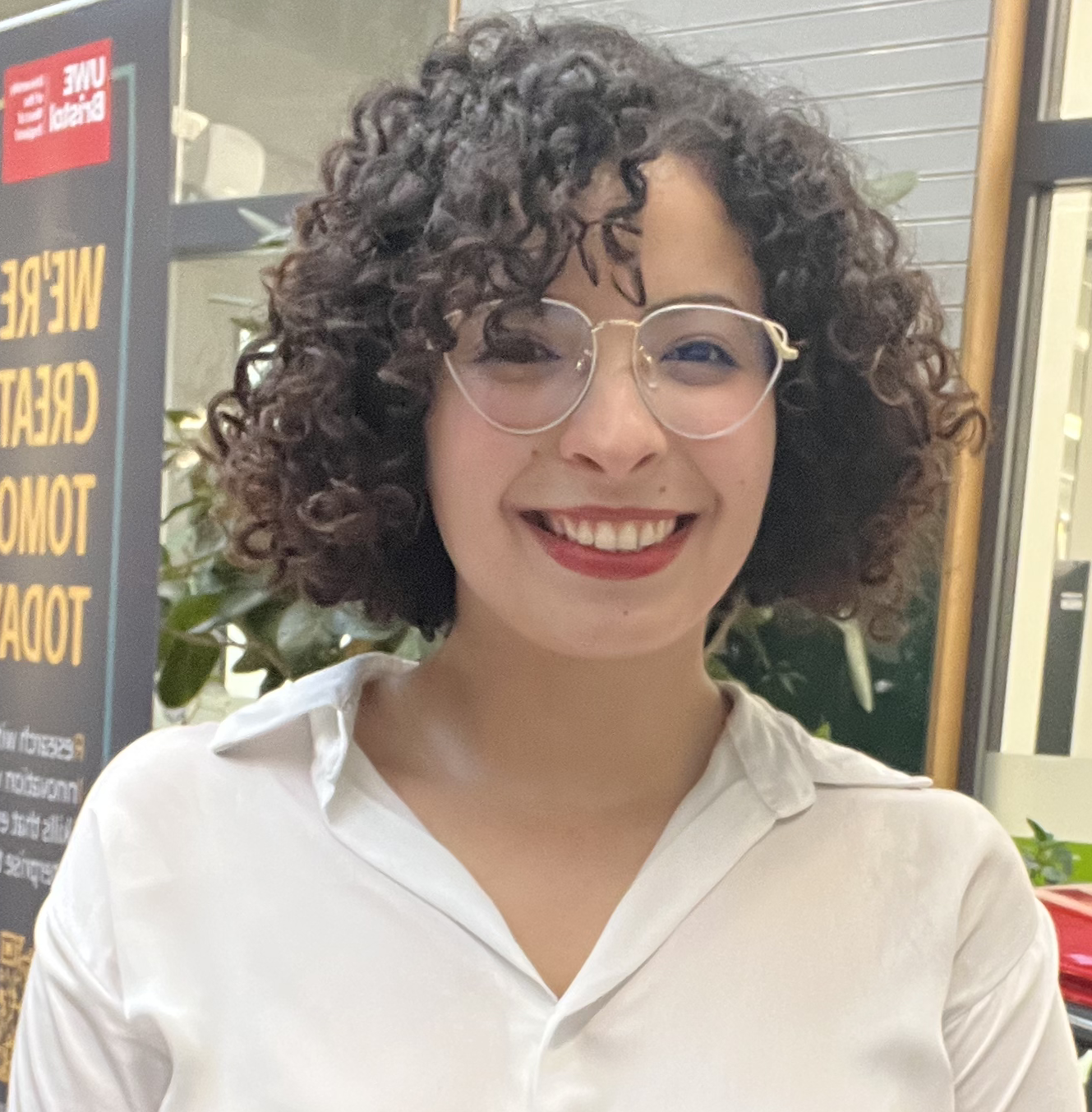}}]
{Dalila Khettaf}~is currently a PhD student at the University of the West of England (UWE) Bristol, where her research lies at the intersection of artificial intelligence and cybersecurity, with a particular focus on collective anomaly detection. She has authored several research papers on this topic, published in respected journals and international conferences. In September 2024, she joined the University of the West of England as an Associate Lecturer, where she teaches a range of undergraduate and postgraduate modules.
\end{IEEEbiography}
\begin{IEEEbiography}[{\includegraphics[width=1.1in, height=1.1in]{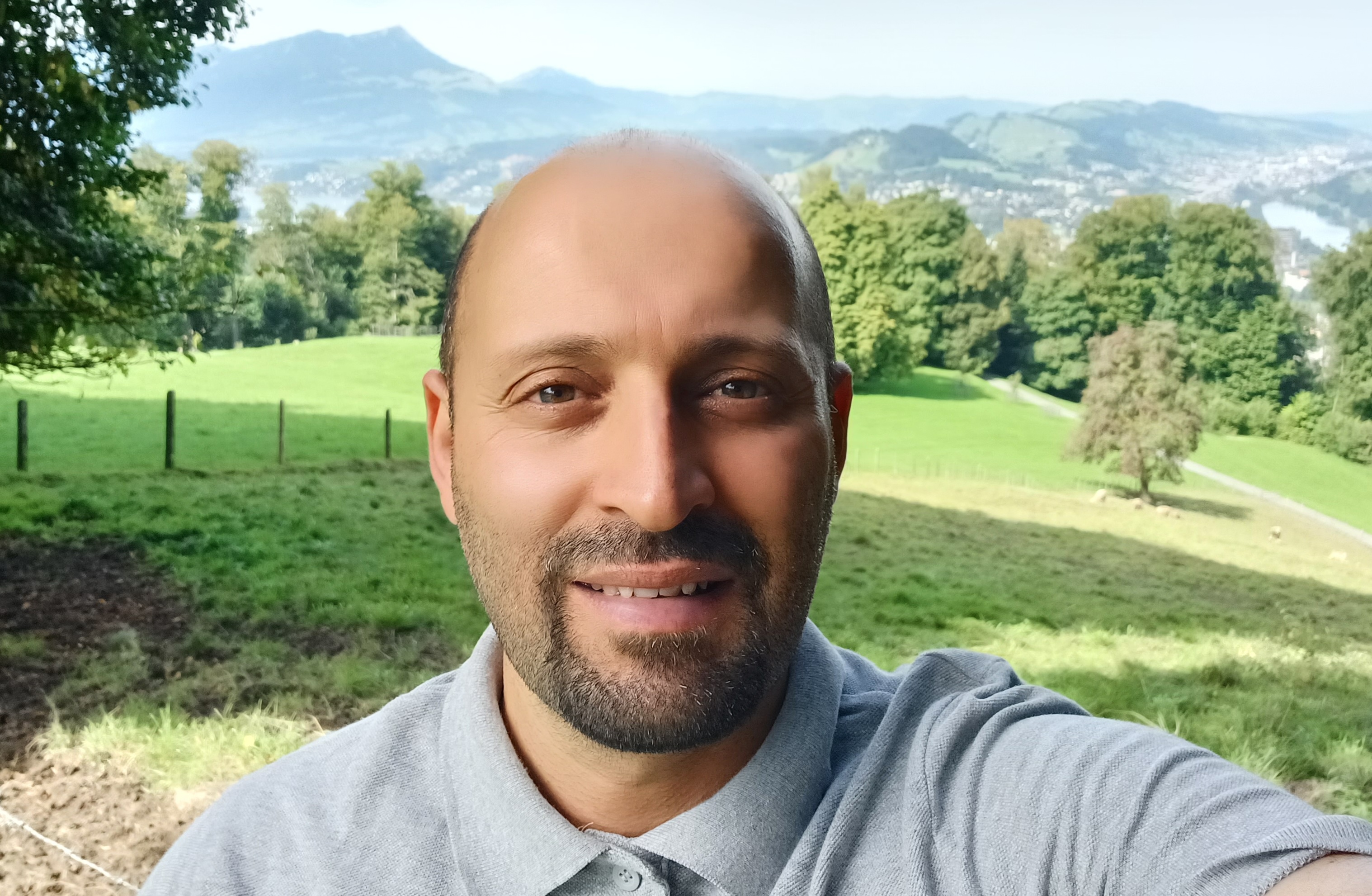}}]
{Djamel Djenouri}~is with the University of the West of England, Bristol, UK, where he is leading many funded research projects. He
obtained a Doctorate in Computer Science then habilitation from the USTHB in 2007 and 2011, respectively. He was an ERCIM postdoctoral fellow at NTNU from 2008 to 2009, then a senior research scientist (Director of Research) and deputy director at the CERIST research center. He also served as an adjunct full professor at Blida University and then at the EMP Polytechnic University. He is being reported amongst the top 2\% most cited scientists in the annual Stanford University releases and ranked in the top 0.5\% of all scholars worldwide by ScholarGPS in his research fields. He is serving as an expert examiner, TPC member of many international conferences, e.g., IEEE ICC, GlobeCom, VTC, ITNAC, WCNC, WiMob, guest editor and a member of the editorial board for many journals such as IEEE Trans. on Sys. Man. and Cybernetics, Future Internet, etc. He has been invited for delivering keynotes, tutorials, and panelist in many international conferences and events. He is a senior member of ACM, IEEE, AGYA Academy Alumni life-member, and a fellow of the UK Higher Education Academy.
\end{IEEEbiography}
\begin{IEEEbiography}[{\includegraphics[width=1in,height=1.25in,clip,keepaspectratio]{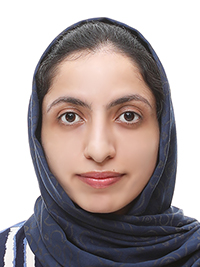}}]
{Zeinab Rezaeifar}~is a Senior Lecturer at the University of the West of England (UWE), UK. She received her Ph.D. in Computer Science and Engineering from Hanyang University, South Korea, in July 2018. Following her Ph.D., she worked as a Postdoctoral Researcher at the Information System Security Laboratory, Korea University, focusing on network security. Prior to joining UWE, she held a Research Associate position in the School of Computing at Ulster University, UK, where she worked on advanced persistent threat detection. Her research interests include network security, anomaly detection, security challenges in 5G and beyond, security issues in Vehicular Ad Hoc Networks (VANETs), security and privacy in Named Data Networking (NDN), and advanced persistent attack detection and reconstruction.
\end{IEEEbiography}
\begin{IEEEbiography}[{\includegraphics[width=1in,height=1.25in,clip,keepaspectratio]{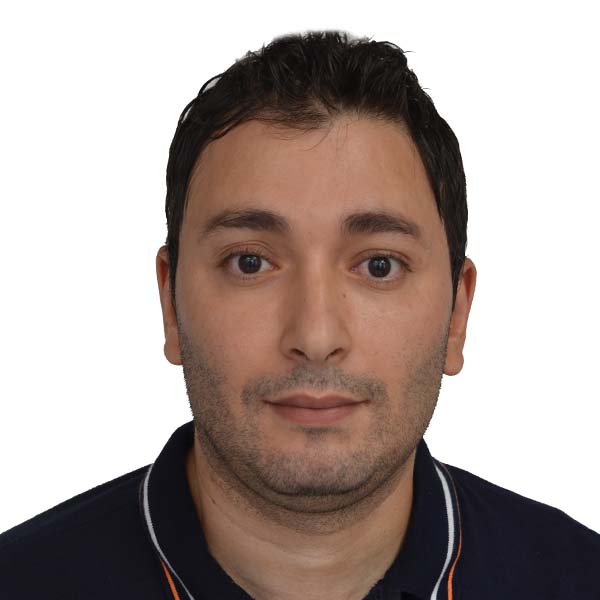}}]
{Youcef Djenouri}~is an Associate Professor at University of South-Eastern Norway, and a senior researcher at NORCE (Norwegian Research Center) from 2023. He was a research scientist at SINTEF, and a postdoc researcher at NTNU, and SDU. His research interests include AI, smart city applications, security and privacy. Dr. Youcef Djenouri published more than 200 research papers in top conferences and journals such as ICDM, ICDE, AAMAS, ACM KDD, IEEE TIST, IEEE TII, IEEE TCYB, and others. He is also in the list of 2\% most outstanding researchers according to Stanford statistics. Dr. Youcef Djenouri is an Associate Editor in IEEE Transactions on Computational Social Systems, Neural Processing Letters, Discover AI journal, and Editorial Board in Applied Intelligence. He also organized workshops and special sessions in top conferences such as ICDM, KDD, DSAA, IJCNN, and PAKDD.
\end{IEEEbiography}
\EOD
\end{document}